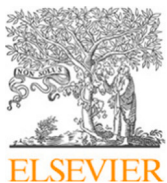

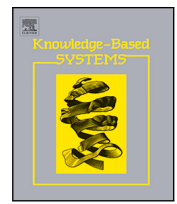

# A comprehensive survey on integrating large language models with knowledge-based methods

Wenli Yang [a],*, Lilian Some [a], Michael Bain [b], Byeong Kang [a]

[a] *University of Tasmania, Churchill Ave, Hobart, 7005, TAS, Australia*
[b] *University of New South Wales, High St, Sydney, 2052, NSW, Australia*

## ARTICLE INFO

## ABSTRACT

The rapid development of artificial intelligence has led to marked progress in the field. One interesting direction for research is whether Large Language Models (LLMs) can be integrated with structured knowledge-based systems. This approach aims to combine the generative language understanding of LLMs and the precise knowledge representation systems by which they are integrated. This article surveys the relationship between LLMs and knowledge bases, looks at how they can be applied in practice, and discusses related technical, operational, and ethical challenges. Utilizing a comprehensive examination of the literature, the study both identifies important issues and assesses existing solutions. It demonstrates the merits of incorporating generative AI into structured knowledge-base systems concerning data contextualization, model accuracy, and utilization of knowledge resources. The findings give a full list of the current situation of research, point out the main gaps, and propose helpful paths to take. These insights contribute to advancing AI technologies and support their practical deployment across various sectors.

## 1. Introduction

The rapid development of Large Language Models (LLMs) has shaped the contours of AI. This has landed the models in areas where they can excel in generating natural text, never reaching any peak before. LLMs are based on deep learning structures, and can therefore perform effectively in many natural language processing tasks, such as text generation, sentiment analysis, or sophisticated dialogue systems.

Recent surveys have explored diverse aspects of LLMs, such as their architectures, training methodologies, and performance evaluation benchmarks. Many focus on specific topics, such as detailed analyses of state-of-the-art models [1], scaling laws innovations((i.e. what a new law brings to make old predictions still work)) [2], and pretraining methods used with large datasets [3]. Others investigate domain-specific fine-tuning effects from reinforcement learning or human feedback [4,5], and transfer learning strategies [6]. Despite these valuable contributions, however, there remains a paucity of comprehensive perspectives that connect the foundational principles of LLMs with practical applications and the challenges faced as these models are implemented in real-world situations.

This survey addresses this gap by presenting a comprehensive analysis of LLMs' foundational principles and their applications across diverse domains. Although LLMs have achieved remarkable progress, their practical deployment faces challenges. Issues such as interpretability, high computational demands, and scalability impede their broader adoption. This study also investigates the integration of generative AI with knowledge bases, emphasizing how this synergy can mitigate these limitations and unlock new opportunities.

To guide this analysis, several key assumptions are outlined:

**Real-World Application Challenges:** LLMs encounter substantial obstacles in real-world settings, particularly in terms of interpretability, computational requirements, and scalability, which limit their effectiveness and broader applicability.

**Mitigation through Integration:** The integration of LLMs with knowledge bases — using methods such as Retrieval-Augmented Generation (RAG), Knowledge Graphs, and Prompt Engineering — offers promising solutions. This synergy enhances data contextualization, improves model accuracy, and reduces computational costs.

**Barriers to Adoption:** Persistent challenges, including the need for interpretability, efficient resource utilization, and seamless integration with existing systems, continue to hinder the widespread adoption of LLMs.

Building on these assumptions, this survey provides a structured and integrated analysis of LLMs. The primary contributions are as follows:

---

* Corresponding author.
*E-mail address:* yang.wenli@utas.edu.au (W. Yang).

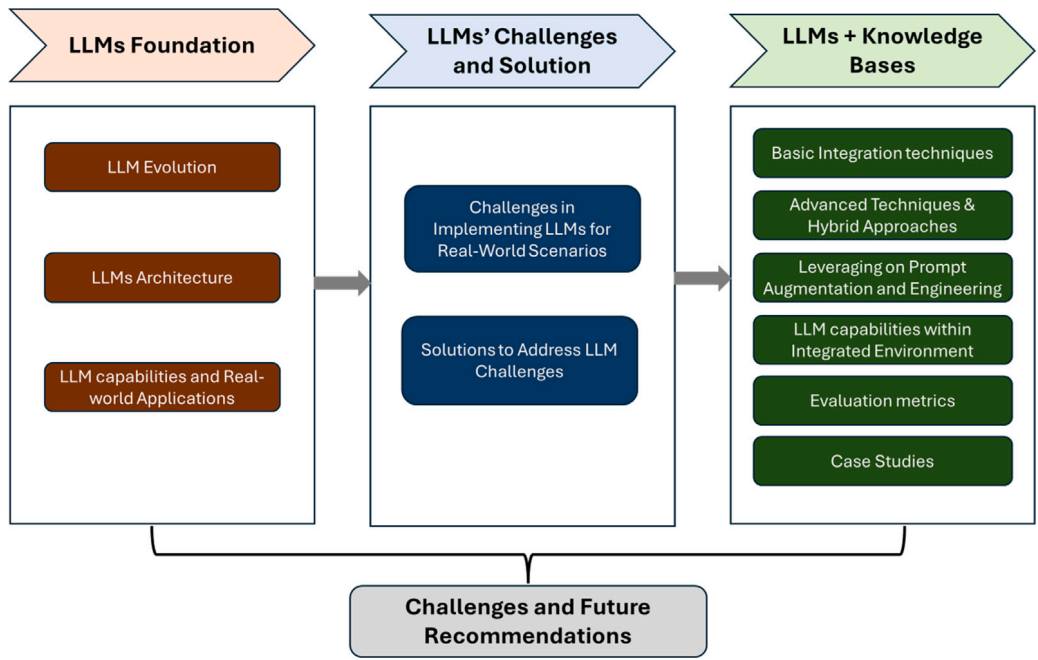

**Fig. 1.** Overall structure of the paper organization.

- Providing a comprehensive analysis of LLMs, focusing on their foundational principles, architectures, and real-world applications, while highlighting their capabilities, as well as technical, operational, and ethical limitations, and their transformative impact across diverse domains.
- Investigating the integration of LLMs with knowledge-based methods, analyzing their technical capabilities, hybrid integration strategies, and real-world case studies, while summarizing evaluation metrics used to assess the effectiveness of LLM-knowledge integration.
- Critically evaluating the challenges of LLM-knowledge integration and presenting actionable recommendations to enhance efficiency, interpretability, and scalability in future AI research and applications.

This survey paper begins with an overview of Large Language Models (LLMs), detailing their evolution, underlying architecture, and diverse real-world applications. It then explores the challenges associated with implementing these models, categorizing them into technical, operational, and ethical/social dimensions and discussing potential solutions. Building on this foundation, the Integrating LLMs with Knowledge Bases section examines integration techniques designed to enhance LLMs with structured knowledge, advanced hybrid approaches and prompt augmentation strategies. This section also provides practical insights through case studies, illustrating real-world applications and evaluation metrics for knowledge-integrated LLMs. Finally, the paper discusses the challenges of integration and offers recommendations for future development and implementation, providing actionable guidance to advance the field. Fig. 1 shows the overall structure of the paper organization.

## 2. Overview of LLMs

### 2.1. LLM evolution

Large Language Models (LLMs) are advanced AI systems capable of understanding and generating human-like text by leveraging deep learning architectures trained on vast amounts of textual data. Built primarily on the transformer architecture, LLMs power applications in machine translation, question-answering, dialogue systems, and generative AI [7–9]. Fig. 2 visually represents the progression of LLMs over time, highlighting key milestones and advancements in AI language processing. The development of LLMs can be categorized into three major phases:

**Early NLP Models (Pre-2010s):** Early natural language processing (NLP) models relied on rule-based and statistical methods, focusing on predefined patterns and probability-based predictions. Statistical NLP techniques, such as n-grams, TF-IDF (Term Frequency-Inverse Document Frequency), and Naïve Bayes, analyzed word frequency and co-occurrence patterns but lacked contextual understanding and semantic depth [10]. As the field progressed, Neural NLP (NLMs) introduced word embeddings like Word2Vec and GloVe, which captured word relationships in continuous vector spaces, significantly enhancing representation learning [11]. The development of Recurrent Neural Networks (RNNs), Long Short-Term Memory networks (LSTMs), and Gated Recurrent Units (GRUs) further improved sequential data modeling, allowing for better context retention. However, these models still struggled with long-range dependencies, limiting their effectiveness in processing complex linguistic structures [12].

**Transformer Era (2017–2022):** The introduction of scalable, self-attention-based models revolutionized natural language processing (NLP) by replacing Recurrent Neural Networks (RNNs) with the self-attention mechanism, enabling parallel computation and improved scalability [13]. This breakthrough led to the development of Pre-Trained Language Models (PLMs), such as BERT, which introduced bidirectional encoding, and GPT, which leveraged auto-regressive generation, setting new benchmarks for NLP performance across various tasks [7]. As LLMs scaled, GPT-3 demonstrated zero-shot and few-shot learning capabilities, showcasing the potential of large-scale models in handling diverse tasks with minimal fine-tuning. However, this rapid expansion also introduced challenges such as bias, factual inaccuracies, and high computational costs, necessitating further advancements in model efficiency and knowledge augmentation [14].

**Next-Gen LLMs (2023–Present):** Recent advancements in Large Language Models (LLMs) have been guided by four interrelated priorities: multimodal capabilities, computational efficiency, knowledge augmentation, and advanced reasoning abilities. These directions aim to improve the performance, usability, and reliability of LLMs in increasingly complex and diverse applications. First, in the area of multimodal capabilities, models such as GPT-4 and Google Gemini represent significant progress by incorporating multiple modalities — including text, images, and audio — into a unified framework. This integration enables richer contextual understanding and supports more advanced reasoning across domains involving heterogeneous data sources [15]. Second, computational efficiency and accessibility have been key drivers in the development of open-source LLMs. DeepSeek LLM [16] exemplifies this trend, offering a scalable and transparent alternative to proprietary models. Trained on high-quality multilingual and code datasets, it demonstrates competitive performance across a range of NLP tasks. Within the DeepSeek ecosystem, DeepSeek-Coder [17] addresses the growing demand for efficient code generation, while DeepSeek-MoE [18] introduces a Mixture-of-Experts (MoE) architecture that improves inference efficiency by activating only a subset of model parameters during prediction. Similarly, LLaMA 3 (Meta AI) and Mistral 7B prioritize lightweight model design and training optimization, enabling faster deployment and wider accessibility for both academic and industrial applications [19]. Third, growing interest in knowledge augmentation reflects the need to overcome LLM limitations in factual accuracy and domain-specific reasoning. Hybrid approaches such as Retrieval-Augmented Generation (RAG), Graph-Enhanced LLMs (e.g., GraphRAG, KAG) [20], and LLM-driven reasoning frameworks (e.g., ToG, KGAgent) [21] integrate structured knowledge sources, including knowledge graphs and document corpora, to ground model outputs and reduce hallucinations [15,22]. Finally, advanced reasoning abilities have emerged as a distinct focus, emphasizing the development of models capable of logical inference, multi-step reasoning, and dynamic problem-solving. A representative example is OpenAI-o1 model [23], released in late 2024, which introduces a reasoning-optimized architecture. By dynamically allocating more computational resources to tasks of higher complexity, o1 improves performance in areas that demand deeper cognitive processing and task-aware inference.

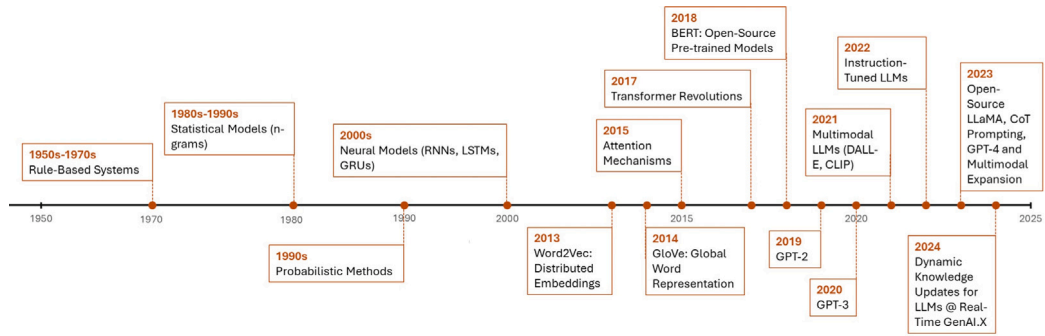

**Fig. 2.** The evolution of large language models.

### 2.2. LLM architecture

Modern LLMs are primarily based on the transformer architecture, which can be easily adapted for specific tasks with robust performance, leading to large gains on downstream tasks like text classification, language understanding, reference resolution, common-sense inference, summarization, and machine translation [7]. Building upon the transformer, LLMs are typically designed using one of three main architectures: encoder, decoder and encoder–decoder [24,25]. The Table 1 summarized these three main architectural design, highlighting their examples, primary use cases, and knowledge integration benefits.

**Encoder LLMs**: These models, such as BERT, RoBERTa, focus on text classification and retrieval tasks. They process input text bi-directionally, allowing them to generate high-quality embeddings that improve retrieval-augmented systems. By learning contextual relationships, encoder models enhance semantic search, information extraction, and retrieval-based NLP applications.

**Decoder LLMs**: Examples include GPT-4, LLaMA-3, DeepSeek LLM, and OpenAI o1, which are mainly designed for tasks such as text generation, dialogue systems, code synthesis, and creative writing. These models generate text auto-regressively, predicting each token based on prior context. DeepSeek LLM exemplifies efficient open-source performance across both understanding and generation tasks, while OpenAI o1 introduces a novel reasoning-optimized architecture that differentiates it from conventional decoder models. Instead of allocating a fixed amount of computational resources uniformly across all inputs, o1 dynamically adjusts its internal computation in response to task complexity. Although highly capable in generative tasks, decoder models often benefit from retrieval-augmented generation (RAG), which grounds outputs in external knowledge to reduce hallucinations and improve factual accuracy in knowledge-intensive domains.

**Encoder-Decoder LLMs**: Models like T5 and BART leverage both an encoder and a decoder to transform input text into output sequences, making them well-suited for machine translation, text summarization, and paraphrasing. Their dual-component structure allows them to integrate effectively with structured knowledge bases (KBs) and semantic reasoning systems, enhancing their ability to provide factually accurate and contextually relevant outputs.

Furthermore, specialized transformers such as Reformer, Longformer, and knowledge-enhanced LLMs address specific challenges in efficiency, sequence length processing, and factual grounding. Reformer reduces memory and computational costs by leveraging locality-sensitive hashing and reversible layers, enabling efficient handling of long sequences [34]. Similarly, Longformer optimizes attention mechanisms to improve context retention in lengthy documents, allowing tokens to focus on globally significant and locally relevant information [35]. The evolution of LLMs has also seen a shift towards task-specific, decoder-only models for applications like text generation. Additionally, the rise of open-source models such as LLaMA and BLOOM has driven the democratization of LLM research, fostering transparency, reproducibility, and collaborative development [36]. These open-access models reduce dependence on proprietary systems, support domain-specific adaptations, and promote innovation across academia and industry, ensuring broader accessibility and practical advancements in knowledge-augmented AI [14,37].

### 2.3. LLM capabilities and real-world applications

Large Language Models (LLMs) exhibit a diverse range of capabilities that enable their application across various domains. While their core functionality revolves around text generation, their broader abilities extend to language understanding, reasoning, decision-making, knowledge management, multimodal processing, and adaptability [26, 38,39]. Table 2 presents a structured summary of LLMs' key capabilities, categorized alongside real-world applications and supporting references.

LLMs excel in natural language processing (NLP) tasks such as translation, summarization, and text completion, generating contextually relevant responses [9,40]. They also engage in complex reasoning tasks, including question answering (QA) and logical problem-solving, making them valuable in decision support, scientific research, and automation [41]. Furthermore, LLMs facilitate large-scale information retrieval, integrating knowledge from external sources like ERNIE and E-BERT to enhance their contextual awareness [42].

**Table 1**
Categorization of Large Language Models (LLMs) based on architecture.

| Model type | Examples | Primary use cases | Knowledge integration benefits | Refs |
|---|---|---|---|---|
| Encoder-Only | BERT, RoBERTa, ModernBERT | Text classification, retrieval | Improves embeddings for retrieval-augmented systems | [8,26,27] |
| Decoder-Only | GPT-4, LLaMA-3, DeepSeek LLM, OpenAI-o1 | Text generation, dialogue, creative writing | Can be enhanced with knowledge retrieval techniques like RAG | [15,17,26,28, 29] |
| Encoder-Decoder | T5, BART, mBART, DeltaLM | Machine translation, summarization | Works well with structured KBs and semantic reasoning | [30–33] |

Beyond structured tasks, LLMs contribute to creative domains, generating storytelling, poetry, and other artistic content. Their adaptability is further demonstrated through fine-tuning for domain-specific applications such as financial forecasting, medical diagnosis, and legal analysis. Additionally, multimodal capabilities allow LLMs to process and generate content combining text, images, audio, and video, expanding their applicability to sectors such as entertainment, healthcare, and security [43,44].

Prompting techniques, including chain-of-thought reasoning, enhance LLMs' problem-solving abilities, while in-context learning enables them to adapt to new tasks with minimal supervision [45,46]. Domain-specific models, such as Financial LLMs and Medical LLMs, illustrate the trend towards specialization, allowing for more accurate and efficient task execution in industry-specific scenarios [41]. Furthermore, LLMs are evolving into autonomous agents capable of executing tasks based on natural language instructions, controlling software applications, and integrating with robotics [47].

Despite their vast potential, LLMs face challenges, particularly in specialized domains requiring high precision, such as medicine and law. Their reasoning capabilities, factual consistency, and computational resource demands remain significant hurdles [10,39]. Nevertheless, LLMs continue to revolutionize industries, from policy-making and finance to healthcare and crisis management, transforming how knowledge is processed and applied.

One emerging approach to overcoming these challenges is the development of cross-domain LLMs. While there is no explicit definition of "cross-domain LLMs", discussions around mixed-domain training and the integration of external knowledge bases suggest that these models are designed to function effectively across multiple knowledge domains. Cross-domain LLMs exhibit key characteristics such as knowledge generalization, multitasking, transfer learning, fine-tuning, and scalability [76–78]. Their inherent versatility allows them to be applied across various industries, including legal document analysis, customer service, and academic research. For instance, Financial LLMs (FinLLMs) leverage mixed-domain training on both general and financial corpora, incorporating prompt engineering and instruction fine-tuning to adapt models to financial tasks [59]. Similarly, Medical LLMs (MedLLMs) apply LLM technology to healthcare, utilizing specialized models such as PharmacyGPT and Psy-LLM to enhance domain-specific performance [79].

## 3. Challenges in implementing LLMs for real-world scenarios

Implementing LLMs in real-world applications presents numerous challenges that span technical, operational, and ethical dimensions. Despite the extensive capabilities, LLMs experience limitations in the following areas: accuracy and reliability, explainability, reliance on data lacking updating mechanisms, and broader concerns such as information provenance, privacy, data security, and potential plagiarism. These challenges necessitate ongoing research and development to ensure the responsible and effective use of LLMs in real-world contexts.

### 3.1. Technical challenges

Technical challenges with LLMs include model interpretability, as their complex decision-making processes are difficult to understand, and the need for substantial computational resources, which can be costly and inefficient.

### 3.1.1. Model interpretability

Even though LLM models show remarkable performance across a range of tasks, it is still challenging to comprehend the logic underlying the text predictions or generation. Concerns arise from this lack of openness, particularly in important applications where justifiable decision-making is imminent as often users seek to understand how and why a model is returning specific outputs. Such transparency helps build or create confidence, ensures ethical application, or encourages further innovation on the platform [80]. Other concerns are caused by model complexity explainability in decision making, bias [81], debugging and error identification [82], trust in high-stake use [15] and ethical concerns. More so, LLM models are referred to as 'black boxes' because they are intrinsically complex and difficult to decipher [10,14]. The models frequently have millions to billions of parameters, resulting in a complex network of linked nodes that support generative functionality. Because of this complexity, it is difficult for humans to trace the specific model's inputs and their outputs [77]. An example of such a problem is GPT-3, which has 175 billion parameters. LLMs are trained using deep learning methods, often called "black-box" models, which means that even experts cannot fully access or understand their internal mechanisms. This lack of transparency hinders efforts to comprehend how particular inputs lead to specific outcomes [14]. LLM interpretability poses challenges to trust and adoption. Thus, in many real-world use cases, it would be advantageous for the stakeholders to understand and justify what the LLMs are doing, especially in areas that are strictly governed by laws, such as medicine, finance, and legal systems [83]. It is equally challenging but important to ensure that LLMs are not perpetuating prejudices or rendering unjust or discriminatory outcomes if we do not comprehend the LLM decision-making process.

### 3.1.2. Managing computational resources

The development of LLMs remains overly complex and needs robust computational power for both training and inference operations. The associated computational costs in LLMs range from training cost, inference time, resource restraints in IDE, and token costs in conversation-style APR, prompting length costs to optimization of multiple sub-tasks. Training big language models is an expensive undertaking that calls for a lot of computer power. For instance, 64 32-GB V100 GPUs were utilized in training Microsoft's InferFix model [15]. This underscores the necessity of significant hardware investments, which may be a challenge for investors and institutions with constrained budgets. Sources cite that energy consumption during training is a growing concern due to the carbon footprint being generated by the AI industry, raising environmental sustainability issues [84–86]. Additionally, longer inference times can cause delays in response generation, even when larger LLM's hardware is well equipped [87]. In real-time applications like chatbots, virtual assistants and interactive systems where prompt replies are essential, this could be troublesome.

**Table 2**
Categorization of key capabilities of LLMs.

| Category | Key capabilities | Applications | Refs |
|---|---|---|---|
| Language understanding and generation | – Natural language processing (NLP): Translation, summarization, text completion. | Generating legal, medical, and business documents; automating content for journalism, marketing, and education. | [48–50] |
| | – Creative content generation: Storytelling, poetry, conversations. | | |
| | – Document creation and automated drafting. | | |
| Reasoning and decision-making | – Complex reasoning: Question answering (QA), logical reasoning, problem-solving. | Medical diagnostics, fraud detection, legal case evaluation. | [12,29,38,45, 51–53] |
| | – Decision support: Data-driven insights, strategic planning, risk assessment. | | |
| Knowledge management and retrieval | – Information retrieval: Data extraction, synthesis, summarization. | Medical knowledge extraction, legal case research, AI-powered chatbots, enhancing search engines. | [54–59] |
| | – Knowledge integration: Leveraging external knowledge bases (e.g., ERNIE, E-BERT). | | |
| Code & program development | – Code generation: Writing, debugging, optimizing | Intelligent script generation, task automation. | [47,60,61] |
| | – Automation: Generating scripts, optimizing workflows. | | |
| Multimodal processing | – Multimodal integration: Processing and generating outputs combining text, images, audio, and video. | AI-powered radiology image analysis, video summarization for media, education, and security monitoring. | [62–64] |
| Adaptability and learning | – Prompt engineering: Few-shot and zero-shot learning, chain-of-thought reasoning. | Financial trend analysis, supply chain optimization, adaptive tutoring systems in education. | [38,53,59,65, 66] |
| | – Domain specialization: Tailoring models for industries (Finance LLMs, Medical LLMs.) | | |
| Autonomous functionality | – Autonomous agents: Interacting with environments, planning, executing tasks. | AI-driven robotic process automation, AI-powered customer service agents. | [67–71] |
| Sectoral impact and creativity | – Cross-industry applications: Transforming healthcare, education, finance, research. | Ideation in creative arts, scientific innovation, entertainment. | [72–75] |
| | – Creative problem-solving: Generating solutions for ill-defined challenges. | | |

Moreover, LLMs require substantial storage and runtime memory, and integrating them into Integrated Development Environments (IDEs) can be difficult, posing challenges for code completion, bug-detection and code generation [88]. Performance problems may arise from coding LLMs, particularly on devices with limited resources. This demand for high computational power can limit their use, especially for small organizations [89]. In addition, the cost of generated tokens and input can add up in conversation-style Automated Program Repair (APR), where LLMs work together iteratively to produce fixes. This can result in considerable costs [15].

### 3.2. Operational challenges

Operational challenges are equally significant, particularly concerning the integration of LLMs with existing infrastructures and ensuring scalability and reliability; maintenance and updates are prevalent across domains.

#### 3.2.1. Integration with existing systems

The LLM integration process is not simply about connecting software or systems; the process requires a multifaceted approach that considers numerous factors, including architectural compatibility, data flow management, and operational efficiency.LLMs are typically based on intricate deep learning architectures, which may not easily interface with systems that may use outdated technologies or have distinct architectural designs. For instance, integrating with legacy systems that do not have access to contemporary APIs or that use different data formats can be quite difficult [90].

Concurrently, during the integration process, data flow management between LLMs and systems is vital for data consistency [91–93]. Research reiterates the significance of improving data access and suggests time-and-space efficient methods for LLMs as it impacts cost, latency, and hardware strain because integration processes require a large amount of processing power [41,92]. Nevertheless, it may not always be possible to effectively optimize operational LLMs for specific scenarios since it demands a significant amount of computational resources. Therefore, it is essential to lessen the pre-training load and increase retrieval efficiency to improve operational efficiency in production-ready systems [94].

### 3.2.2. Scalability and reliability

Scalability refers to an LLM system's capacity to handle increasing workloads and data volumes while maintaining performance, whereas reliability concerns the system's ability to consistently produce accurate and trustworthy outputs, regardless of the operating environment or task complexity [95]. Research investigates computational demands, data complexity sensibility to model fine-tuning and adaptability as the major concerns for scaling LLMs. Scalability is a salient feature exhibited by LLMs that enables them to effectively manage workload increase, which becomes challenging because of the enormous amount of processing power needed for both training and inference. Maintaining performance while scaling becomes more difficult as the model size and data volume increase. This frequently leads to increased expenses, resource usage, and possible processing bottlenecks [96,97]. For example, to ensure that jobs are equitably dispersed over various servers and that network latencies are kept to a minimum when growing LLMs across distributed infrastructures, complex orchestration is needed. The difficulty of guaranteeing reliability — the system's capacity to operate consistently under a range of loads and conditions — also confronts large-scale deployments. Due to LLM's intricate structures, reliance on knowledge bases, and hardware constraints, LLMs are prone to malfunctions [7]. LLM systems must be created to function at various levels of throughput and be optimal in any operating environment. This is particularly critical in industries where availability and uptime are critical, such as the computer, automation, manufacturing, navigation, and software industries [89]. Striking a balance between data inputs involves multiple trial-and-error attempts, making the fine-tuning process difficult. This unpredictable nature of model training underscores the need for flexible LLM infrastructure and subsequently the computing power [98]. Multi-bucketing and continuous batching techniques optimize the computational efficiency of LLMs by selecting appropriate buckets for data input and enhance multi-text request processing thus, using computational resources efficiently. These problems are mitigated by constructing a perspective and robust infrastructure, incorporating redundancy, continuous management, and iterative improvement of the system. Robust error handling, redundancy methods, and system monitoring are strategies that provide continuous availability and minimize downtime in high-demand systems. Additionally, upholding scalability and reliability calls for eliminating failure points, and preventing service degradation to support LLM functions in a production environment.

### 3.2.3. Maintenance and updates challenges

Large language models (LLMs) present maintenance and updating difficulties that must be resolved to maintain system stability and dependability. To evolve and address ongoing LLM problems, models undergo continual updates. However, ensuring compatibility with new versions, managing model updates without disrupting existing workflows, and addressing security vulnerabilities require robust maintenance procedures [93]. Robust maintenance procedures are necessary to provide compatibility with new versions, manage upgrades without interfering with existing operations, and address security risks as models undergo regular modifications to adapt and handle ongoing LLM problems [83,93,99]. Managing numerous versions of LLMs requires efficient version control and rollback methods that enable prompt recovery in the event of malfunctions by rolling back to earlier versions. Furthermore, to avoid unanticipated problems or performance degradation and to make sure that any changes enhance the system without creating new problems, comprehensive testing and validation of updates before deployment are essential. Both the ongoing system performance monitoring and the gathering of user input preceding LLM updates are crucial because they reveal hidden issues and potential areas for development and allow for prompt adjustments to increase reliability. Adaptability is an essential feature that new LLM versions must incorporate for LLMs to stay useful and efficient in a quickly changing technical environment [29]. Finally, to prevent creating vulnerabilities or jeopardizing sensitive data during updates/upgrades, security and privacy issues must be properly mitigated. Strong security protocols are essential to safeguard information security and maintain user trust during the update process.

### 3.3. Ethical and social implications

LLMs are plagued by multiple ethical and social problems related to bias perpetuation, misinformation, the need for transparency in AI-driven decision-making, and potential impacts on employment. A primary concern is whether the algorithm operates fairly and responsibly, given the general risk that the technology can be misused [80,83]. Additionally, the credibility of sources daunts the LLMs family, as the models do not trace the provenance of the content they generate, creating uncertainty around source reliability [100].

The reliance on proprietary datasets such as those used in OpenAI poses risks to privacy and data security. In some instances, LLMs may produce data similar to those from training data without attribution, causing ripples in creative and academic spheres as it raises concerns about plagiarism and intellectual property [29,101]. Financial LLMs, for example, face unique challenges in ensuring data privacy while utilizing proprietary data without breaches [12]. Moreover, the automation of mundane tasks by LLMs could impact employment, displacing roles traditionally performed by humans.

Addressing these social issues calls for the development of new ethical standards within the generative AI sphere and the implementation of measures to reduce bias and embrace transparency in Gen AI operations [102]. Techniques like RAG are increasingly adopted in financial domains to address these privacy and trust challenges. By tackling these ethical and social issues, stakeholders can work to ensure the responsible use of LLMs [101].

In conclusion, LLMs are capable of extending new opportunities for innovative development, but it is necessary to resolve the associated challenges. In this way, focusing on the improvement of the technical characteristics of the LLM systems and the operational issues — such as interpretability, efficiency, integration, and scalability, ensures that the LLM technologies are applied appropriately and responsibly [89]. Emergent and evolving challenges and broader concerns necessitate adaptive strategies involving regulatory frameworks, ethical guidelines, and technical solutions like differential privacy and federated learning [38]. Such approaches assist in developing trustworthy and accurate LLM systems that will be useful in generating reliable outcomes in various applications [103].

## 4. Solutions to address LLM challenges

As large language models (LLMs) continue to face significant challenges in real-world applications, addressing these issues is essential for ensuring their responsible and effective deployment. Key challenges such as interpretability, computational resource demands, and data quality require targeted solutions. Enhanced data training methods are crucial to enhance the accuracy and relevance of LLM outputs, by ensuring these models are trained on diverse, high-quality datasets. To tackle the challenge of computational inefficiency, the development of efficient algorithms and advanced hardware is necessary, enabling LLMs to operate at scale without excessive resource consumption. Explainable AI (XAI) techniques will address concerns about model interpretability, helping users understand the reasoning behind LLM outputs and increasing trust in these systems. Additionally, addressing privacy, data security, and ethical concerns requires the integration of regulatory frameworks, ethical guidelines, and technical solutions like differential privacy and federated learning. By focusing on these solutions, the challenges surrounding LLMs can be mitigated, ensuring that these technologies can generate reliable, ethical, and efficient outcomes across diverse sectors.

### 4.0.1. Enhanced data training techniques

Research stresses that high-quality training data is the fundamental component of robust and effective LLMs because extensive text datasets teach LLM patterns, trends, and other insights. However, incomplete, biased, or out-of-date data will be passed over and retained thus compromising the resulting LLM [90,104]. Domain-specific fine-tuning improves LLMs by training them on specialized datasets catered to specific industries, such as banking, biomedicine, and medical education [12,105,106]. In finance, fine-tuning can be done with news articles, market data, and company financials to enhance their ability to perform financial forecasting, risk analysis and investment analysis [12]. This increases LLM proficiency with domain-specific tasks by acquainting them with pertinent terminology and information. Data cleaning and filtering address problems with noisy or biased data, guaranteeing high-quality inputs that prevent models from learning undesirable patterns. Likewise, data augmentation creates new data to expand training datasets when available data is scarce, improving the model's robustness [7].

Integrating several data sources, including text, code, and transcripts of conversations, enhances an LLM's comprehension across different knowledge areas [87,107]. For example, an LLM integrated with data from GitHub, text from Wikipedia and chats from chatbots can perform a wide range of tasks including code generation, text summary and dialogue. However, more research is needed to refine methods for integrating these sources effectively.

Scalability of data curation remains challenging, requiring strong infrastructure and efficient data handling to support storage, indexing, updating, and retrieval of large datasets [95]. As such, robust infrastructure is needed to handle the storage, processing, and analysis of vast data, while efficient handling of data is a prerequisite for indexing, updating and retrieval of heterogeneous data. Given that quality datasets are necessary for the best possible LLM performance, measuring and ensuring data quality is also critical. To fully utilize the promise of these models, research and innovation in scalable data curation and data training are required, a continual action that is critical for enhancing the accuracy, reliability, and fairness of LLMs.

### 4.0.2. Optimizing algorithms and hardware for efficiency

To overcome computational cost challenges, optimized algorithms and hardware should be implemented for improved performance and efficiency.

Research highlights large computational demands placed on Large Language Models (LLMs) during deployment and training, and suggests ways to increase effectiveness. By selectively activating specific parts of the network to optimize resources, techniques such as mixture-of-experts explained (MoE) models can improve performance while decreasing complexity. Distributed computing Frameworks can be adopted for distributing the processing load among several machines and accelerating training [93,108]. More so, deep learning operations have been accelerated by specialized hardware, such as GPUs, and TPUs and further innovation on the hardware will keep computing costs down. Originally intended for rendering graphics in video games and visual effects industries, their parallel processing efficiently, has made GPUs extremely effective in deep learning applications. On the other hand, TPUs which are tailored specifically for deep learning applications, offer an even greater level of computational power and efficiency than GPUs, providing a significant advantage in performance for such tasks [12,36]. Additionally, model compression techniques—such as knowledge distillation and model pruning—enhance deployability on devices with limited resources without compromising performance [95]. Pruning eliminates crucial connections or neurons within the model, while quantization, reduces the precision of numerical values, and knowledge distillation, to replicate the performance of a larger "teacher" model in a smaller 'student' model [29]. Similarly, Parameter-Efficient Fine-tuning (PEFT) like LOoRA reduce processing needs by focusing on fine-tuning a limited set of parameters [14], thus reducing computational overhead. This strategy proves beneficial in financial models to improve effectiveness and flexibility for financial functions where tasks necessitate minor adjustments to specific model parameters rather than a complete overhaul [109]. PEFT can boost the efficiency and adaptability of retrieval models in recommendation systems that cater to changing user behaviors [29] Using distributed computing systems improves productivity by facilitating model training and inference on numerous devices simultaneously. Expanding the workload among multiple sources can greatly diminish the time needed for training and inference processes. This enables the creation and implementation of extensive and intricate models. Examples of frameworks leveraging distributed systems include DenseX, EAR, UPRISE, RAST, Self-MEM, FLARE, Filter-rerank, R-GQA, LLM-R, LM-Indexer, and BEQUE, each optimizing efficiency in large-scale deep learning [77,107]. Finally, developing expansive hardware infrastructure, cloud technology offers scalable, on-demand computing capabilities that enable LLM training and implementation at a reasonable cost.

### 4.0.3. Explainable AI (XAI) for transparency and trust

The key obstacles to making LLM explicable include limited context window, implicit bias, lack of transparency, model complexity and scale, data complexity, and assessment complexity. Despite this, there is considerable research interest in this area to unravel LLMs explainability.

Models have a restricted limited context windows, as a result, LLM limits the input and output of text LLM generates. This constraint hinders the explainability of the process, particularly in summarization tasks, where LLMS struggle due to limited text access [26,87]. Implicit biases from training data can also surface in outputs, making it hard to trace why a model generates biased responses. The bias can be subtle and hard to decipher. For example, gender bias can be exhibited when a nuanced news article is used as part of a training dataset. The internal logic of LLMs is also often opaque, with their complex neural networks hindering transparency and trust in their decisions. As models grow larger, with billions of parameters, explaining their behavior becomes increasingly challenging. The diversity and scale of training data make it hard to identify specific influences on outputs, and the lack of standardized metrics complicates efforts to assess and compare explainability across different LLMs [14].

There are several methods or approaches to improve LLM explainability. Engineering prompts elicit more explicable responses from LLMs through the use of strategies such as interaction and iteration, formatting with an example, explicit instruction, system-specific instruction, and control tokens. An illustration of this might be an explainability prompt like "Explain the concept of LLM as if I am 10 years old" [104,110]. Model interpretation tools like benchmark tests and tools like RGB, RECALL, and CRUD [110,111] and reference tools like model agents like AutoGPT, Toolformer, and Graph-Toolformer where LLMs employ active judgment in their operations can be used to gain insight into the decision-making process of LLMs [104,105,110,112]. Based on the particular requirements of each task, meta-reasoning prompting (MRP) assists LLMs in dynamically choosing and utilizing various reasoning strategies, potentially increasing the transparency of their reasoning process [107].

## 5. Integrating LLMs with knowledge bases

This section explores various techniques for integrating Large Language Models (LLMs) with knowledge bases, highlighting the potential solutions these integrations offer. By examining the methods used to combine LLMs with structured knowledge, we discuss how this synergy enhances the models' ability to provide more accurate, contextually relevant, and data-driven outputs. The integration strategies also open doors to improved reasoning capabilities, allowing LLMs to access and leverage external knowledge to address domain-specific queries and tasks more effective.

### 5.1. Integration techniques for enhancing LLMs with structured knowledge

Techniques for integrating LLMs with knowledge bases aim to enhance the models' performance by grounding their outputs in structured information. Key methods include using knowledge bases (KBs), which provide structured data for LLMs to query, and knowledge graphs, which organize information in interconnected nodes and relationships to improve context understanding. Additionally, Retrieval-Augmented Generation (RAG) combines document retrieval and generation, enabling LLMs to access relevant data from external sources and generate more accurate, up-to-date responses. These integration techniques enhance LLMs' ability to produce reliable, context-aware outputs.

#### 5.1.1. Basic knowledge bases

Basic Knowledge Bases (KBs) are structured repositories that store facts in a relational format, providing an organized way to query and retrieve predefined facts, relationships, rules, and context, offering high precision when factual data is required.

In their inherent nature, Knowledge Bases help improve factual accuracy as they store verified information. Integrating LLM with the Knowledge base enables the models to cross-reference outputs with factual data stored in KBs thereby reducing hallucination and improving reliability. LLMs that access Wikidata, an open data knowledge base storing structured data on various domains (about people, places, and events), can generate factual outputs such as birthdate or scientific facts [87]. Studies show that Wikidata integration with Knowledge Enhanced PLMs promotes entity-aware training, through entity linking to Wikipedia, and joint training based on WiKidata's Knowledge Graph [87]. Several notable models, including K-BERT, KgPLM, FaE, JAKET, LUKE, WKLM, and CoLAKE, utilize Wikidata as a knowledge source in various ways including knowledge injection, knowledge-guided attention, and Knowledge Graph embedding, which improves factual accuracy, reasoning and enhanced interpretability [87]. Specialized knowledge bases like UMLS (Unified Medical Language System) for medical terminology or Freebase for general knowledge, provide LLMs with domain-specific data, enhancing their performance on tasks requiring specialized expertise [87]. For instance, LLMs integrated with the UMLS can retrieve up-to-date information on drug interactions or disease symptoms, which would significantly improve the model's ability to provide medical diagnoses or treatment recommendations [113]. However, most KBs are static and riddled with limited reasoning capacities, often requiring manual upgrades. This makes them unsuitable for real-time or fast-changing information and limits their utility in fields such as current events or rapidly evolving research domains. While KBs store factual data, they do not inherently capture complex relationships or provide the inference capabilities of more advanced systems like Knowledge Graphs. For applications requiring real-time updates or complex inference, knowledge bases alone are insufficient, and more advanced systems are needed.

#### 5.1.2. Knowledge graphs

Knowledge Graphs (KGs) provide a structured approach to representing knowledge, enabling LLMs to overcome inherent limitations in understanding and reasoning about the connections between real-world entities and concepts. This integration is valuable in complex reasoning tasks like multi-hop question answering and semantic disambiguation where connection between concepts is valuable. Additionally, the KGs allow LLMs access to up-to-date knowledge without the need for retraining, making them a powerful tool to keep LLMs relevant and current [114].

Knowledge Graphs (KGs) are integrated into Large Language Models (LLMs) using several advanced methods designed to enhance their reasoning capabilities: embedding, Graph Neural Networks, prompting and semi-structured chain of thought. These techniques vary in approaches to processing and utilizing structured knowledge.

Knowledge Graph Embedding transforms entities and relationships into numerical vectors for efficient processing, while Graph Neural Networks capture complex relationships within the KG, enhancing multi-hop reasoning abilities [115]. Knowledge Graph prompting injects relevant KG data directly into the LLM's input, guiding reasoning and improving accuracy [8]. Semi-structured chain of thought combines both structured KG data and unstructured text to generate reasoning chains, leveraging all available knowledge for more complex reasoning tasks [94,104,115]. Each method provides a distinct way to enhance the LLM's reasoning and response capabilities. Key techniques for integration include KG-enhanced LLMs, LLM-augmented KGs and synergized LLMs and KGs. In KG-Enhanced LLMs, KGs are used to augment the training or inference process of LLMs, supplying them with structured information to support reasoning and understanding. Whereas in LLM-augmented KGs, LLMs assist in various KG-related tasks, such as embedding, completion, construction, and question answering [116]. Lastly, in synergized LLMs and KGs, the unified framework combines the strengths of both KGs and LLMs, creating a model that leverages KGs for structured knowledge and LLMs for language understanding [117].

Notable advantages of Knowledge Graphs over knowledge bases include increased reasoning, inference, and improved interpretability. KGs store information in a network of nodes (entities)and edges (relationships) and emphasize relationships between entities, making them more flexible and dynamic than knowledge bases with unstructured text. Unlike KBs, which store isolated facts, integrated LLMs can reason about complex relationships and infer new information based on existing graphs. As such Knowledge Graphs such as Google Knowledge Graph, ConceptNet and DBPedia, LLMs can answer questions by traversing relationships between entities. Specifically, ConceptNet supports common-sense reasoning, DBPedia provides structured data for general knowledge queries, and Google Knowledge Graph aids in answering factual questions by linking related entities [87,104].

Research demonstrates several ways in which Knowledge Graphs (KGs) can be used to enhance the capabilities of large language models (LLMs). KGs naturally support explainability by visualizing the graph of entities and relationships. KGs can enhance explainability by showing how the model arrived at a specific conclusion. A KG can map the relationships between legal precedents, statutes, and case law, providing an explainable pathway that LLMs follow to generate legal opinions [7,10,114].

KGs can be updated with new information, making it ideal for real-time applications (like stock exchanges, active medical diagnosis, and navigation) where information must be frequently refreshed. Applicable frameworks like MedGraph, WeKnow-RAG and Think on Graph enable LLMs can retrieve real-time knowledge about evolving topics, such as the latest news developments or ongoing live events [76,81,83, 109].

KGs offer a structured and interconnected way to represent knowledge, which is beneficial for LLMs in understanding relationships and making inferences. Unlike relational databases or NoSQL solutions, KGs use graph structure to explicitly link information, making it easier for LLMs to understand relationships and make inferences [36,104,115]. This interconnected structure facilitates the use of specialized query languages (like SPARQL and Cypher) that are well-suited for navigating complex relationships between entities [92,114].

KGs enhance factual accuracy and knowledge probing by providing LLMs with a factual grounding that might be missing or weakly represented in their training data. This is especially helpful for tasks such as answering questions, where providing accurate information is crucial [85]. For example, the LAMA (Language Model Analysis) dataset is designed to test how well LLMs have internalized factual knowledge and by incorporating knowledge from KGs, LLMs can perform better on such knowledge probing tasks [85]. KGs that encode commonsense knowledge, like ConceptNet and ATOMIC, are invaluable for LLMs, they boost common-sense reasoning and natural language generation [85,87].

While KGs offer significant advantages, the tradeoffs include incompleteness, complexity, scalability, and performance. KGs are often plagued with missing links and relationships between entities. LLMs can be trained to predict these missing links, effectively completing and refining the KG, thus benefiting both LLMs, by providing them with a more complete knowledge base, and KGs, by improving their accuracy and coverage [104]. Creating and maintaining large-scale KGs is a resource-intensive process that requires careful design and data curation [95]. Scaling across broad domains or diverse industries can be difficult and costly. Although KGs improve reasoning capabilities, integration with LLMs may introduce latency and reduce overall system performance. Hence, though Knowledge Graphs may offer a more advanced solution than KBs by providing a dynamic, relationship-focused structure that enhances reasoning and interpretability, the complexity and scalability challenges make KGs less feasible for broad, real-time applications unless coupled with significant computational resources. Therefore, KG is best suited for high-stakes domain-specific tasks where reasoning and relationships between entities are critical, such as pharmaceutical drug discovery and research, Failure Mode and Effects Analysis (FMEA), financial fraud detection, and support threat intelligence in cyber-security [114,115,118].

### 5.1.3. Retrieval augmented generation

Retrieval-augmented generation (RAG) is a hybrid approach that integrates LLMs with a retrieval mechanism to fetch relevant information from external documents or knowledge in real-time during the text generation process. This technique allows LLMs to dynamically access vast external corpora (like knowledge graphs, databases or search engines) to retrieve the most relevant information [41,90,105,107]. RAG enables language models to retrieve factual information and generate more accurate and contextually aware outputs, especially in cases where a language model's training data may be outdated or incomplete [99]. This approach provides the LLM with an "external memory" to supplement its internal knowledge base, thus enhancing LLM quality and accuracy.

RAG comprise of retrievers, generators, and knowledge bases [41]. The retriever dynamically fetches relevant information from external corpora, the generator uses this retrieved information to generate a response, and the knowledge base is a collection of text such as scientific articles, news articles, or Knowledge Graphs [95,119,120]. This architecture allows the LLM to access up-to-date knowledge beyond its static training data, alleviating hallucination issues by grounding the generated response in factual data.

RAG enhances retrieval and generation through several key techniques which may include document chunking, embedding models, retrieval techniques, querying, knowledge graph integration, iterative retrieval and generation and self-reflection. Chunking breaks down large text into manageable sizes using a mix of static and semantic methods to maintain context [119,121]. Helps embedding models to represent information and queries in a way that maintains semantic meaning, improving retrieval accuracy [25]. RAG employs retrieval methods such as dense retrieval, which uses vector representations (cosine similarity) for semantic matching, and sparse encoding for keyword matches between queries and documents; with hybrid approaches combining the strengths of both [95,119]. Dense Passage Retrieval (DPR) technique utilizes dense embeddings to match semantically relevant document chunks to queries and the combination of DPR with traditional sparse retrieval (i.e., BM25) has been shown to further enhance retrieval precision in complex or high-precision tasks [82,109].

Query expansion using techniques like Query2doc involves expanding the original query to include additional terms, increasing the effectiveness of retrieval [122]. Furthermore, Knowledge Graphs can be integrated into RAG systems for structured reasoning, allowing more accurate and meaningful results [76,114]. Iterative retrieval and generation involve repeating these processes to refine outputs, guided by self-evaluation mechanisms that assess the adequacy of retrieved information [25,123]. Self-reflection features in advanced systems like Self-RAG allow the model to assess the relevance and accuracy of the information it retrieves and generates, improving overall output quality and contributing to better explainability [105].

Retrieval-augmented generation (RAG) excels in complex, knowledge-driven tasks by linking large language models (LLMs) with real-time or domain-specific retrieval, significantly enhancing the scope and accuracy of AI applications. RAG capabilities are experienced in varied settings. In cross-domain applicability, RAG is used for question answering, dialogue generation, summarization, fact-checking or verification, information extraction, and reasoning [25,36,78,82,114].

In question-answering (QA) systems, Naive RAG improves response accuracy, especially for multi-hop queries or long-form answers where standalone LLMs might lack context. This is crucial for fact-checking, where retrieving authoritative sources ensures the reliability of generated content [35,120]. RAG's role in dialogue systems is equally transformative, as it enriches real-time responses, especially in task-oriented conversations like customer support [124]. Modular RAG capabilities such as text summarization are invaluable for condensing large documents into coherent, concise summaries, boosting content generation efficiency [125]. Additionally, RAG's ability to retrieve and integrate specialized knowledge makes it indispensable in fields such as law and medicine, where precision is critical [126].

Furthermore, RAG strengthens recommendation systems by providing personalized suggestions based on user data, and in code search and generation, it supports developers by retrieving relevant code snippets to address complex technical challenges. Overall, RAG's versatility across diverse applications highlights its vital role in enhancing LLMs' performance, making them more accurate and contextually aware in real-world scenarios.

Overall, RAG systems are ideal for real-time data retrieval of unstructured data in environments with dynamic data volumes. For example, OpenAI RAG offers LLM the ability to query data and retrieve and utilize vast external data in its response, which improves the relevance and factual accuracy of generated text and real-time knowledge access [120]. Therefore, a knowledge graph is best suited for high-stakes, domain-specific tasks where reasoning and relationships between entities are critical. RAG renders domain flexibility through retrieving domain-specific documents from external sources making it easier for LLMs to answer specialized queries without needing extensive domain-specific training. Table 3 summarized techniques for integrating Large Language Models (LLMs) with knowledge bases (KBs), Knowledge Graphs (KGs), and Retrieval-Augmented Generation (RAG).

### 5.2. Advanced techniques and hybrid approaches

The integration of Retrieval-Augmented Generation (RAG) with large language models (LLMs) has significantly improved their ability to handle complex tasks by incorporating external knowledge sources.

**Table 3**
Integration techniques for enhancing LLMs with structured knowledge.

| Technique | Description | Typical methods | Advantages | Challenges |
|---|---|---|---|---|
| Basic Knowledge Bases (KBs) | Structured repositories storing predefined facts, relationships, and rules. | Knowledge Injection, Knowledge-guided Attention, Knowledge Graph Embedding | High factual accuracy, reduces hallucinations, domain-specific knowledge (e.g., UMLS, Wikidata) | Static, limited reasoning, slow updates, lacks complex relationship handling. |
| Knowledge Graphs (KGs) | Represent knowledge through interconnected nodes (entities) and edges (relationships). | Graph Neural Networks, Knowledge Graph Embedding, KG Prompting, Semi-structured Chain of Thought | Enhanced reasoning, dynamic updates, supports multi-hop reasoning, explainability, improves interpretability | Missing links, scalability issues, complexity, resource-intensive maintenance. |
| Retrieval-Augmented Generation (RAG) | Combines document retrieval with generation, allowing LLMs to access external data in real-time. | Document Chunking, Embedding Models, Dense/Sparse Retrieval, Knowledge Graph Integration, Self-Reflection | Real-time access to up-to-date information, dynamic context, alleviates hallucination, accurate outputs. | Requires efficient retrieval systems, latency, computational complexity. |

While traditional RAG enhances factual grounding, advanced techniques extend beyond simple retrieval by incorporating structured knowledge graphs, adaptive reasoning frameworks, and self-optimizing learning mechanisms. This section categorizes LLM enhancement techniques into three progressive levels. Table 4 summarizes the three levels of advanced integration. Each level highlights specific categories, key applications, and associated typical models, demonstrating how knowledge integration enhances LLMs for improved retrieval, reasoning, decision-making, and domain-specific optimization.

- **Level 1: Knowledge Retrieval & Integration** focuses on how LLMs acquire and integrate external knowledge, including structured, unstructured, and multi-modal data.
- **Level 2: Knowledge Utilization & Reasoning** covers how LLMs process, synthesize, and apply retrieved knowledge to produce logically consistent and explainable responses.
- **Level 3: Optimization & Specialization**: explores adaptive learning techniques, reinforcement-based retrieval refinement, and domain-specific customizations to continuously improve LLM decision-making.

### 5.2.1. Knowledge retrieval and integration

At this level, the primary focus is on equipping LLMs with the ability to efficiently retrieve and integrate structured, unstructured, and multi-modal knowledge sources. This foundational stage ensures that LLMs can access relevant, up-to-date, and contextually meaningful information while reducing hallucination risks.

The first category, "Advanced Knowledge Acquisition Strategies", emphasizes developing techniques that allow LLMs to search, filter, and extract information from massive datasets, ensuring high-quality knowledge retrieval. By incorporating retrieval-augmented generation (RAG) techniques and context-aware indexing, LLMs can perform enterprise search, biomedical knowledge retrieval, and open-domain question answering (QA) more effectively. Models such as T-RAG [127] and REALM [82] demonstrate how intelligent retrieval can enhance LLMs' ability to provide reliable, up-to-date responses.

The second category, "Structured Knowledge Graph Integration", further refines retrieval by leveraging knowledge graphs (KGs) and hierarchical structures to organize and enhance reasoning efficiency. By reducing retrieval noise and improving factual consistency, this approach enables LLMs to support domain-specific QA and more structured information synthesis. Key models such as TRACE [123], GraphRAG [128] exemplify how structured knowledge enhances explainability and contextual reasoning. TRACE improves knowledge tracing and retrieval alignment by dynamically filtering and structuring relevant information, reducing retrieval noise, and ensuring contextually appropriate knowledge utilization. GraphRAG extends traditional retrieval-augmented generation (RAG) approaches by leveraging graph-based knowledge representations, enabling multi-hop inference, entity disambiguation, and hierarchical knowledge synthesis, which are critical for high-stakes applications such as biomedical QA and legal AI.

### 5.2.2. Knowledge utilization and reasoning

Building upon retrieval capabilities, Level 2 shifts the focus from knowledge access to knowledge reasoning, inference, and multi-hop processing. LLMs at this stage are expected to go beyond simply retrieving facts and instead synthesize, analyze, and logically structure information for complex decision-making.

The first category, "Contextual Knowledge Reasoning and Inference", enhances LLMs' ability to process and connect multi-hop information across diverse data sources. This allows for advanced reasoning tasks such as multi-hop QA, legal AI, and decision support systems. LLMs must reduce logical inconsistencies and enhance interpretability to support users in making informed decisions. For example, FlexKBQA [129] utilizes KG embeddings to improve entity linking and retrieval efficiency. ToG [130] enables LLMs to perform multi-hop reasoning over knowledge graphs, improving structured inference and knowledge grounding. Additionally, models such as ReAct [131], and Think-on-Graph 2.0 [132], Chain-of-knowledge [133] focus on structured inference and enhanced logical synthesis, further strengthening LLMs' ability to generate well-reasoned and contextually accurate responses.

The second category, "Adaptive Knowledge Infusion for Decision-Making", ensures that LLMs can dynamically ground their reasoning in real-world data, selectively incorporating relevant information based on the task's needs. This is particularly critical for interactive AI agents and high-stakes applications such as medical decision-making, financial analysis, and crisis response systems. A notable example is MetaRAG, which integrates retrieval-augmented generation with metacognitive capabilities to enhance multi-hop question-answering tasks. It applies metacognitive principles — monitoring, evaluating, and planning — enabling LLMs to self-assess their reasoning paths and dynamically adjust based on identified errors or knowledge gaps [41].

### 5.2.3. Optimization and specialization

At the highest level of integration, LLMs are optimized for long-term performance, self-improving retrieval strategies, and domain-specific knowledge refinement. This stage ensures that models are highly precise, reliable, and adaptable for critical industries such as healthcare, legal compliance, and scientific research.

The first category, "Self-Optimizing Knowledge Systems", enables LLMs to continuously learn and adapt through reinforcement learning, retrieval fine-tuning, and real-time system updates. By self-improving over time, these models enhance their reasoning capabilities while minimizing errors in high-stakes applications. Key techniques such as CRAG [36] and WeKnow-RAG [134] employ adaptive retrieval enhancement and self-supervised fine-tuning, ensuring that LLMs dynamically refine their knowledge-grounding processes. Similarly, KGA-gent [135] enables LLMs to autonomously explore, retrieve, and infer structured knowledge from knowledge graphs, providing real-time reasoning adjustments and optimized knowledge integration. Additionally, Coarse-to-Fine Highlighting [136] improves retrieval granularity and hierarchical knowledge filtering, allowing AI models to prioritize and contextualize retrieved information more effectively.

**Table 4**
Three levels of advanced integration.

| Level | Category | Key applications | Description & benefits | Key models |
|---|---|---|---|---|
| Level 1: Knowledge retrieval & integration | Advanced knowledge acquisition Strategies | Enterprise search, biomedical retrieval, open-domain QA. | Focuses on retrieving and integrating structured, unstructured, and multi-modal knowledge sources. Improved retrieval accuracy, better contextualization, reduced hallucination risks. | T-RAG, REALM, DPR-RAG |
| | Structured Knowledge Graph Integration | Domain-specific QA. | Enhances LLMs by incorporating Knowledge Graphs (KGs) and hierarchical reasoning structures to improve retrieval efficiency and factual accuracy. Reduced noise in retrieval, enhanced explainability, structured reasoning. | Triple-aware reasoning, TRACE, GraphRAG |
| Level 2: Knowledge utilization & reasoning | Contextual knowledge reasoning & Inference | Multi-hop question-answering, legal AI, decision support systems. | Allows LLMs to process and synthesize multi-hop knowledge across structured and unstructured data sources for complex reasoning. Enhanced interpretability, reduced logical inconsistencies, multi-step reasoning. | FlexKBQA, ToG, ReAct, KG-RAG4SM, SPINACH, Think-on-Graph 2.0, Chain-of-Knowledge |
| | Adaptive Knowledge Infusion for Decision-Making | Interactive agents, high-stakes applications. | Enables dynamic knowledge grounding where LLMs selectively incorporate relevant information based on task-specific needs. Improved trustworthiness, verifiable knowledge sources, more reliable decision-making. | MetaRAG, MoE-RAG KAG |
| Level 3: Optimization & specialization | Self-optimizing knowledge systems | Autonomous systems, real-time adaptation, AI assistants. | Enhances LLMs through adaptive learning, reinforcement learning, and self-improving retrieval strategies. Continuous improvement in retrieval and reasoning, long-term optimization. | FiD-RAG, WeKnow-RAG, KGAgent, FUNNELRAG |
| | Domain-specific knowledge customization | Healthcare, legal compliance, scientific research. | Tailors LLM performance for domain-specific needs, ensuring precision and reliability in industry applications. Highly specialized LLMs, increased factual reliability, optimized retrieval in constrained domains. | Self-BioRAG, MedGraphRAG, SAC-KG |

The second category, "Domain-Specific Knowledge Customization", focuses on fine-tuning LLMs for specialized applications that demand precision, compliance, and factual reliability. This is particularly crucial in healthcare, legal AI, and scientific research, where incorrect or imprecise knowledge can lead to significant consequences. Traditional fine-tuning approaches rely on static knowledge integration, but recent advancements in Knowledge-Augmented Generation (KAG) [137] have introduced adaptive knowledge grounding, enabling LLMs to dynamically incorporate, verify, and refine domain-specific knowledge during inference. Models such as Self-BioRAG [105], MedGraphRAG [121], and SAC-KG [138] exemplify how domain-specialized expertise can be embedded and dynamically retrieved within LLMs. Self-BioRAG improves biomedical knowledge synthesis by integrating structured retrieval mechanisms with adaptive fine-tuning, ensuring high accuracy in medical decision-making. MedGraphRAG enhances medical AI applications by leveraging graph-based knowledge augmentation, allowing

LLMs to navigate complex, multi-relational medical datasets for improved clinical reasoning and diagnostic accuracy. Meanwhile, SAC-KG optimizes legal and regulatory compliance AI, ensuring that retrieved knowledge aligns with jurisdictional regulations, ethical guidelines, and factual accuracy requirements.

### 5.3. Leveraging on prompt augmentation and engineering for LLM-knowledge base integration

This section introduces foundational prompt engineering techniques designed to enhance the retrieval and generation capabilities of large language models (LLMs). By strategically adapting prompts, these techniques guide models to produce more accurate, contextually relevant, and domain-specific outputs. This foundational approach sets the stage for the subsequent exploration of advanced integration methods, focusing on optimizing the interaction between LLMs and knowledge bases for improved performance across various applications.

Chain-of-thought (CoT) prompting has emerged as a powerful technique for guiding LLMs in complex reasoning tasks by encouraging them to embrace step-by-step reasoning pathways, before formulating a response. CoT exhibits human-like patterns in solving multi-step problems such as arithmetic problems, commonsense reasoning, and symbolic tasks. [83,90,107,110].

Building upon CoT, several enhancements, including Buffer of Thoughts (BoT), Strategic Chain-of-Thought (SCoT), and Graph of Thought have been developed. BoT introduces a meta-buffer that stores high-level thought templates distilled from previous problem-solving processes. These templates can be retrieved and instantiated for new tasks, enabling more efficient and accurate reasoning. BoT outperforms standard multi-query prompting methods by achieving superior performance in tasks requiring multi-step reasoning, such as mathematical problem-solving and logical reasoning, while reducing computational costs. The use of a meta-buffer allows for the creation of generalizable and reusable reasoning structures, optimizing both reasoning accuracy and efficiency. This approach mirrors human cognitive processes of using "mental templates" or heuristics to solve problems [112]. Strategic Chain-of-Thought (SCoT) enhances CoT by incorporating strategy elicitation to guide reasoning processes. It employs a two-stage process within a single prompt, enhancing the stability and consistency of reasoning paths [122]. SCoT outperforms existing methods like Self-consistency and Buffer of Thoughts by using strategic knowledge to generate accurate single-query reasoning paths, demonstrating the importance of strategic planning in LLM reasoning. Both BoT and SCoT leverage cognitive-inspired methods to refine reasoning paths, demonstrating a trend towards incorporating human-like cognitive processes, such as strategy selection and problem decomposition, into LLMs.

Graph-of-Thoughts (GoT) technique extends CoT by representing the reasoning process as a graph, allowing for more complex and non-linear reasoning patterns. This approach enables the exploration of multiple reasoning paths and the evaluation of different options, potentially leading to more accurate and insightful solutions [91,97, 112]. Other chains of thought methods include Mindmap, IRCOT, Reasoning on Graphs, CoT with Consistency, Program Aided Language Model (PAL), Reason and Act, reflection and Tree of thought [40,41, 43,45,115]

Advanced prompting augmented methods are critical for maximizing LLM performance across various tasks. This is exhibited in prompt augmentation systems (PAS), and Slim Proxy Language (SlimPLM) alongside Self-memory techniques. The Prompt Augmentation System (PAS) introduces a novel and data-efficient method for enhancing prompts. As a plug-and-play automatic prompt engineering system, generating complementary prompts to enhance LLM outputs. This method aligns with high data efficiency, achieving state-of-the-art results with minimal data. PAS significantly outperforms Automatic Prompt Engineering (APE) models, demonstrating the effectiveness of generating high-quality complementary prompts automatically. The emphasis on data efficiency and automatic augmentation aligns with broader trends in developing scalable and adaptable prompt engineering solutions for LLMs. It also underscores the importance of reducing the dependency on large datasets, making LLM applications more accessible and sustainable [97].

SlimPLM (Slim Proxy Language Model), is a smaller proxy language model that assesses the LLM's knowledge and determines whether retrieval is necessary, optimizing the use of external resources [139]. Self-memory prompting is an adaptive form of prompting where the LLM's own generated outputs are used to enrich subsequent prompts through self-feedback loops [78], hence improving the quality and consistency of responses. This prospective prompting framework solves the limitations of using the internal memory of traditional RAG by incorporating a retrieval-augmented generator and a memory selector. The generator uses both the input text and retrieved memory to generate output, while the memory selector refines this output to create an unbounded memory pool that is iteratively used for subsequent generations with quality improvement in tasks such as Neural Machine Translation, abstractive text summarization, and dialogue generation [78].

In summary, prompt engineering techniques play a crucial role in enhancing the accuracy, efficiency, and reasoning capabilities of large language models (LLMs). By refining how prompts are structured, these techniques optimize retrieval and generation processes, enabling LLMs to perform more effectively across diverse tasks. They serve as a foundational element for developing robust, context-aware AI applications, laying the groundwork for more sophisticated and adaptable models in real-world scenarios.

### 5.4. LLM capabilities within integrated environment

This section focuses on how integrated large language models (LLMs) effectively address the challenges encountered by traditional LLMs, as discussed in Chapter 2. By leveraging different integration techniques, such as the incorporation of knowledge graphs, knowledge bases, and specialized retrieval mechanisms, these integrated models overcome limitations in accuracy, scalability, and interpretability, providing more robust and contextually aware solutions. The Table 5 highlights the features of different methods for integrating knowledge with Large Language Models (LLMs), focusing on key aspects such as data structure, accuracy, reasoning ability, and application areas.

### 5.5. Evaluation metrics for knowledge-integrated LLMs

#### 5.5.1. Enhancing interpretability in integrated LLMs

Retrieval-augmented generation (RAG) serves as a bridge between the generation and retrieval processes, allowing users to directly trace the pathways of information used by the LLM. By integrating retrieval-based methods, RAG systems give LLMs real-time access to external, up-to-date data sources, such as large document databases, the web, or specialized knowledge bases [36,120]. This provides anchoring for the model's generated responses, enhancing interpretability and transparency by allowing users to verify the source of the information. It also reduces the risk of hallucinations, a common issue with LLMs that rely on static training data [90].

RAG is especially effective for tasks that require current or specific knowledge. For instance, in domains such as biomedicine or law, where verifying conclusions against well-established knowledge is essential, RAG ensures that the LLM's responses are grounded in authoritative, real-world data [105]. The Chronicles of RAG has demonstrated how effective external knowledge integration can be improving trust and transparency by allowing users to track the origin of the generated content [119].

Knowledge Graphs (KGs) provide a structured representation of relationships between entities. By allowing LLMs to access well-organized, structured information, KGs enable LLMs to reason through complex tasks in a logical and traceable manner [36]. This enhances model transparency, as users can follow the reasoning process more clearly. For instance, systems like TRACE use KGs to map logical connections between retrieved evidence, enabling multi-hop reasoning (i.e., connecting multiple sources to answer complex queries) [123]. The integration of KGs into LLMs thus, strengthens the interpretability and traceability of their outputs.

Advanced prompting methods, such as Chain-of-Thought (CoT) prompting, further improve interpretability by breaking down complex questions into smaller, more manageable sub-tasks. This step-by-step reasoning enables LLMs to adopt intermediate reasoning that can be traced and verified, making the decision-making process clearer to the user [83,124].

Systems like TRACE construct reasoning chains that decompose a task into sub-steps, guiding the LLM through each sub-task by identifying key pieces of evidence and logically connecting them. CoT is particularly useful for tasks that require logical reasoning or multi-step calculations, such as mathematical problem-solving, multi-hop question answering, or decision-making tasks [36].

**Table 5**
Comparative analysis of integrating LLMs with knowledge bases.

| Feature | Knowledge Bases (KB) | Knowledge Graphs (KGs) | Retrieval-Augmented Generation (RAG) | Prompting techniques |
|---|---|---|---|---|
| Primary benefit | Provides static, factual data with high precision and reliability. | Enables complex reasoning and inference, emphasizing entity relationships. | Allows real-time retrieval and contextual response generation. | Enhances model response quality through structured prompts, improving clarity and relevance. |
| Data structure | Structured relational format storing facts, rules, and relationships. | Graph-based structure with nodes (entities) and edges (relationships) for interconnected knowledge. | Flexible, utilizes external data sources for real-time retrieval and integration into outputs. | No structured data; relies on pre-trained model knowledge, augmented by prompt patterns and templates. |
| Applications | Best suited for fields requiring reliable, factual information (e.g., scientific data retrieval, expert systems, question answering, semantic search). | Ideal for applications requiring dynamic knowledge updates, such as recommendation systems, drug discovery, knowledge discovery, fraud detection. | Useful in dynamic scenarios needing up-to-date information, like customer support, content generation, question answering (single-hop/multi-hop), information extraction, dialogue generation, code search, text generation, text summarization, text classification, sentiment analysis, and math problem solving. | Effective in tasks requiring nuanced language, creativity, or simulated problem-solving, such as text generation, code generation, translation, summarization, question answering, reasoning, and chatbots. |
| Computational efficiency | Relatively high efficiency; queries are fast due to structured data. | Moderate; graph traversal is efficient but can be complex in large KGs. Querying for complex relationships can be computationally intensive. | Varies; efficiency depends on retrieval model complexity and data volume. | Highly efficient; relies on model's internal processing without external retrievals, minimizing latency. |
| Response accuracy | High accuracy for factual data; may struggle with queries requiring inference or reasoning beyond the explicitly stored knowledge. | High for structured and relationship-driven data, depending on KG completeness and correctness. | Moderate to high; enhanced by real-time retrieval from verified sources. | Variable; accuracy depends on prompt design and model's internal knowledge, which may be outdated. |
| Adaptability to different data types | Low adaptability; generally static data and hard to update. | Moderate; adaptable within domain-specific contexts and relationships. | High; adapts well across domains by retrieving context-specific information. | Very adaptable for structured text tasks but limited to model's pre-trained knowledge without updates. |
| Domain-specific knowledge suitability | Effective for fields with specialized, static knowledge requirements (e.g., medical terminology databases). | Beneficial for domains needing advanced contextual relationships (e.g., legal databases, scientific research). | Highly adaptable; ideal for any domain needing real-time, responsive data integration. | Limited by the scope of model training data; requires carefully crafted prompts for niche topics. |
| Strength in reasoning | Limited; relies on static data and predefined relationships. | Strong, supports multi-hop reasoning and can infer relationships between concepts. | Limited to contextual retrieval; depends on external sources for expanded reasoning. | Provides indirect reasoning via prompt chaining and contextual prompts but lacks true inference ability. |
| Update capability | Static, requiring manual updates for new data. | Flexible, can be updated frequently for real-time data (e.g., live news, ongoing medical developments). | Real-time access to external sources ensures content is current; continuous updates required for relevance. | No direct update capability; prompts use model's training data, which may be outdated. |
| Challenges | Limited scalability without regular updates; static data may impact performance in dynamic applications. | Data consistency issues in large-scale graphs, with a need for frequent updates. | Latency and complexity in real-time retrieval, leading to performance trade-offs under high load. | Risk of generating incorrect or irrelevant responses if prompts are ambiguous or complex; lacks external grounding. |

When combined, RAG, Knowledge Graphs, and Chain-of-Thought prompting create a powerful framework for enhancing interpretability in large language models (LLMs). This integrated approach grounds LLM outputs in verifiable real-world sources reduces hallucinations and offers a structured reasoning framework. Together, these methods increase traceability and make the model's decision-making processes more understandable and easier to validate. This combined framework significantly improves both the reliability and interpretability of LLMs.

### 5.5.2. Managing computational costs in integrated LLMs

Techniques like Retrieval-Augmented Generation (RAG) aim to mitigate these challenges by offloading knowledge retrieval to external databases, reducing the need for models to store all knowledge internally. This external retrieval system allows for more efficient use of resources, both in memory and processing and enables the model to access up-to-date information dynamically. For instance, an exploration of RAG demonstrates how retrieving external documents allows the LLM to reduce its internal storage needs, thus lowering computational demands during both training and inference [119]. This flexibility enhances the scalability of the models, making them more adaptable to real-time updates without requiring retraining, which is resource-intensive [119].

Further improvements to RAG's computational efficiency have been achieved through optimized retrieval methods, such as graph-based re-ranking (G-RAG). G-RAG refines the retrieval process by prioritizing the most relevant documents through a semantic graph that links related documents. This targeted retrieval not only reduces the volume of irrelevant information processed but also increases the precision of the responses, thus reducing overall computational costs [119]. This re-ranking technique significantly minimizes the workload on LLMs while maintaining high response quality. However, while re-ranking shows promise, there is still room for improvement, particularly in reducing the time overhead associated with complex knowledge graph constructions.

Another critical technique contributing to computational efficiency is few-shot prompting, which allows LLMs to adapt to new tasks without requiring full retraining. Few-shot prompting operates by providing a few examples to the model during inference, enabling it to generalize across similar tasks [108,120]. This method, as exemplified in RAG, reduces the need for continuous updates or domain-specific retraining, a process that traditionally demands significant computational resources. The ability of models to adapt dynamically via few-shot learning echoes broader trends in transfer learning, where models pre-trained on large datasets are fine-tuned on smaller, task-specific data. Few-shot prompting, though efficient, may have limitations in complex domains where more extensive task-specific training might be necessary for optimal performance [78,140].

While RAG, optimized retrieval methods, and few-shot prompting present robust solutions for computational efficiency, some challenges remain unaddressed. For example, RAG's reliance on external databases introduces potential delays in retrieval and integration, which may not always align with real-time processing needs. Furthermore, as noted in the critique of RAG, over-reliance on external documents can lead to instances where the model unnecessarily retrieves information it already possesses, contributing to inefficiency. Additionally, research suggests that even with optimization strategies, LLMs sometimes struggle with long or complex queries, such as those affected by the "Lost in the Middle" problem, where the model focuses too much on the start and end of a document while neglecting the middle sections [83,141]

### 5.5.3. Solving scalability and reliability in integrated LLMs

Scaling Large Language Models (LLMs) across various domains presents challenges in terms of handling diverse queries, and vast datasets, and ensuring relevant, accurate responses. A key solution to enhance scalability is to combine an external Retrieval-Augmented Generation (RAG) with Knowledge Graphs (KGs) [41,118,121]. This hybrid system leverages the dynamic knowledge retrieval capabilities of RAG with the structured reasoning of KGs, allowing LLMs to efficiently scale across different domains while maintaining factual accuracy. RAG enables real-time retrieval of external information, hence preventing the need for LLMs to store all knowledge internally. When integrated with KGs, which organize and structure domain-specific data such as medical or legal information, the system ensures precise reasoning and decision-making, especially in large specialized LLM [81,95]. Similarly, in the *Graph-Based Retriever* approach discussed, a hybrid RAG-KG system enhances scalability by addressing the issue of information overload in biomedical literature [81,109,118,142]. In the medical domain, repetitive and redundant data can overwhelm retrieval systems. By using KGs to down-sample over-represented topics and ensure balanced information retrieval, the hybrid RAG-KG system ensures that critical but less prominent information is not overshadowed, making it scalable across large, complex datasets. This hybrid approach broadens the applicability of LLMs to fields requiring precision, such as biomedicine and law, without sacrificing accuracy or efficiency [118].

RAG enhances reliability by grounding responses in external, verifiable sources, rather than relying solely on pre-trained model weights. This grounding ensures that LLM outputs are based on up-to-date, trustworthy information. Structured retrieval methods, particularly graph-based reranking (G-RAG), further improve reliability by optimizing which documents are retrieved [143]. Graph neural networks (GNNs) evaluate the relationships between retrieved documents, elevating the most contextually relevant ones, and reducing the chance of irrelevant or inaccurate information being used in the final output.

In tasks like open-domain question answering, where reasoning often requires drawing from multiple sources, G-RAG's structured retrieval process ensures the model has access to the most reliable documents. Research has demonstrated that by employing Abstract Meaning Representation (AMR) graphs to map semantic relationships between documents, G-RAG filters out less relevant data and prioritizes documents that offer more accurate answers. This significantly reduces the risk of hallucinations — situations where the model generates inaccurate or fictional information — making the system more dependable for open-domain tasks [143]. Lastly, leveraging smaller proxy models to identify knowledge gaps in the LLM, the system can selectively engage retrieval mechanisms, optimizing resource utilization [139].

### 5.5.4. Seamless integration of LLMs with existing systems

Integrating Large Language Models (LLMs) into existing systems presents a variety of challenges, including data access, real-time updates, scalability, security, and interpretability. To address these, multiple advanced techniques have been developed beyond the common methods like Retrieval-Augmented Generation (RAG) and Knowledge Graphs (KGs).

RAG allows LLMs to dynamically retrieve relevant information from external sources during inference, bridging gaps in knowledge and providing access to both structured and unstructured data. LLMs are further enhanced with API integration and data pipelines, ensuring seamless real-time access to diverse data sources such as legacy systems, external APIs, and proprietary databases [41,92]. This integration can be simplified through RESTful APIs or more complex solutions like GraphQL, which allows querying across multiple data endpoints efficiently.

In scenarios where keeping LLMs updated is critical, especially in dynamic fields like healthcare and finance, RAG retrieves the latest information at inference time, eliminating the need for constant retraining. Knowledge Graphs (KGs), which offer structured data and incremental updates, further allow LLMs to access evolving knowledge without undergoing full retraining cycles. This combination is particularly useful in specialized domains where new information is frequently introduced, such as in BIORAG for biomedical data.

Tool-integrated reasoning through search engines, calculators, and code interpreters can extend LLM capabilities, and enhance accuracy, efficiency and integration with specific tasks or domains [29]. Incorporating tools into the thinking process of LLMs can boost effectiveness in tasks that require calculations or specialized algorithms, hence enhancing their performance and lessening their workload. Models such as TORA use tool-integrated reasoning to tackle challenging problems by interweaving natural language reasoning with external tools, like computation libraries and symbolic solvers [144].

Additionally, adapters can be used to fine-tune LLMs for specific tasks without altering the core model parameters [124]. These lightweight neural modules can be plugged into the LLM and trained with domain-specific data, making them highly adaptable to different tasks and reducing computational costs. For example, the UP-RISE framework employs adapters to select the most appropriate prompts from a pool for a specific zero-shot task [36]. This flexibility is particularly useful for enterprise systems, where multiple domain-specific adaptations may be required.

Another promising method for improving scalability is multi-agent architectures, where distinct agents handle specific system tasks, such as data ingestion, knowledge retrieval, and response generation. This division of labor distributes the workload more efficiently and allows LLMs to integrate smoothly with complex systems [145].

To further enhance the system's ability to scale, model merging can combine the parameters of multiple pre-trained models into a unified model, integrating different task-specific capabilities into a single, more versatile LLM. Additionally, tool augmentation extends the LLM abilities by integrating it with external tools, APIs, and real-time data streams for specialized tasks like scheduling, booking, or domain-specific reasoning.

### 5.5.5. Adapting LLMs to changing knowledge bases

One prominent issue is limited temporal knowledge, how models handle time-related information and reasoning. Despite their vast capacity, traditional LLMs are bound by the constraints of their training data. They do not automatically acquire new information after training, which can lead to outdated or irrelevant responses, particularly concerning recent events or discoveries. For example, an LLM trained before a significant geopolitical event or medical breakthrough may still produce content that reflects an outdated understanding. To address this, research explored methods like continuous learning and dynamic training. Continuous learning enables LLMs to incrementally adapt to new information, allowing them to incorporate evolving language and updated knowledge without needing to retrain from scratch [146]. This is particularly valuable in fields where timely and accurate information is crucial. Dynamic training similarly focuses on updating models continuously, ensuring that LLMs remain relevant over time as new research and discoveries emerge [147]. Another substantial challenge is the difficulty in updating knowledge. Traditionally, modifying an LLM's knowledge base requires retraining the entire model, a process that is both computationally intensive and costly [148]. This has spurred the development of techniques like Retrieval-Augmented Generation (RAG), fine-tuning, and incremental learning. RAG systems are particularly effective because they allow LLMs to retrieve external information during inference, circumventing the need for constant retraining [95,109,127].

Through retrieval, models can access up-to-date knowledge from databases, Knowledge Graphs, or other dynamic sources like the Internet. Sources advocate future research on several areas time-aware retrieval, forward-looking active retrieval augmented generation, real-time QA, and timeliness in Gen IR [36,99,147].

Time-aware retrieval potential incorporates temporal metadata like timestamps into the retrieval process of RAG to access the up-to-date data. FLARE anticipates future queries during text generation enabling more timely and relevant access [99]. Realtime QA evaluates the LLM's ability to handle recent events whilst GenIR advocates for research in real-time knowledge access and continual learning and editing to maintain knowledge currency [29]. Notable research is demonstrated with common sense reasoning explored reasoning about time, where LLMs are instructed to retrieve information given a specific time or events [112].

Knowledge-integrated Large Language Models (LLMs) leverage external knowledge sources, such as knowledge graphs (KGs), domain-specific corpora, and databases, to enhance factual accuracy and reasoning capabilities. Evaluating these models requires a structured approach that extends beyond traditional NLP benchmarks, incorporating tailored metrics that assess knowledge retrieval, reasoning, and text generation.

To systematically assess the performance of knowledge-enhanced LLMs, we categorize key evaluation metrics into three main groups:

- Retrieval-based metrics: These metrics assess the efficiency and accuracy of retrieving relevant knowledge from external sources, whether structured (KGs) or unstructured (text-based documents).
- Knowledge consistency and reasoning metrics: These metrics evaluate the model's ability to generate responses that are logically coherent and aligned with factual knowledge.
- Generation and fluency metrics: These metrics measure the overall quality and readability of the generated output, which is crucial for effective knowledge communication.

### 5.5.6. Retrieval-based metrics

Retrieval mechanisms are foundational for knowledge-enhanced LLMs, determining how effectively relevant information is identified and incorporated into responses. These metrics focus on both the accuracy of retrieved content and the efficiency of the retrieval process.

***Recall@K***: Measures the proportion of relevant documents or KG entities retrieved within the top K results [137]. A high *Recall@K* suggests that the model effectively retrieves a significant portion of the relevant knowledge, minimizing the risk of missing critical information. This is particularly important in applications such as medical diagnosis and legal document retrieval, where omitting key details could lead to incomplete or incorrect conclusions.

***Hits@K***: Evaluates whether at least one correct knowledge element appears in the top K retrieved items, ensuring that relevant knowledge is present within the retrieved results [135]. Unlike *Recall@K*, which measures the proportion of relevant items retrieved, *Hits@K* focuses solely on the presence of at least one relevant result, making it highly useful for applications such as question-answering systems, semantic search, and recommendation systems. For example, in chatbot knowledge retrieval, a high *Hits@K* ensures that at least one retrieved document contains useful and factually accurate information, allowing the chatbot to provide meaningful responses.

***Mean Reciprocal Rank (MRR)***: assesses the position of the first correct result in the retrieved list, rewarding models that rank relevant knowledge higher. It is calculated as the average reciprocal rank of the first relevant document across multiple queries, ensuring that knowledge-integrated LLMs prioritize the most relevant information [149]. MRR is particularly crucial in applications where users expect immediate access to useful information, such as search engines and automated legal or financial assistants. For example, in a legal AI system, if a user is searching for case precedents related to intellectual property law, a high MRR ensures that the most relevant case appears at the top of the retrieved results, significantly improving efficiency and decision-making. However, MRR is inherently user-centric, as different users may have varying expectations regarding what constitutes the most relevant result. For example, a legal expert might prioritize precedent-setting cases, while a general user might prefer easily interpretable summaries. This subjectivity means that while MRR is useful for ranking efficiency, it may not fully capture individual user preferences.

***Area Under the Curve (AUC)***:is another important metric for evaluating retrieval performance, particularly in ranking-based retrieval systems. *AUC* measures the model's ability to distinguish between relevant and non-relevant retrieved items, offering an aggregate assessment of ranking quality [138]. A higher *AUC* score indicates that the model consistently ranks relevant documents higher than irrelevant ones, making it a valuable metric in information retrieval, recommendation systems, and search engine optimization. Additionally, *AUC* accuracy, as an extension of general accuracy [130], assesses how effectively a model classifies relevant and irrelevant information, ensuring that the retrieved content aligns with user expectations and query intent.

***Needle-in-a-Haystack (NIH) Test***: evaluates the model's ability to retrieve rare but critical facts hidden within a vast amount of information, making it essential for assessing long-tail knowledge retrieval performance [150]. Many real-world applications, such as biomedical research, financial fraud detection, and cybersecurity threat analysis, require identifying rare but highly significant details that could be easily overlooked. For example, in medical research, a rare but crucial study linking a genetic mutation to a disease may be buried among thousands of more general studies. A knowledge-integrated LLM with strong NIH performance would successfully surface such rare insights, ensuring that vital but infrequent knowledge is not lost. This metric is particularly valuable in high-stakes decision-making environments where missing a rare but critical fact could lead to erroneous conclusions.

### 5.5.7. Knowledge consistency and reasoning metrics

For a knowledge-integrated Large Language Model (LLM), ensuring that the retrieved information is accurately interpreted, logically structured, and coherently applied in its generated responses is just as important as retrieving relevant knowledge itself. A model that retrieves the right information but misinterprets, distorts, or fails to integrate it meaningfully can still produce misleading or incorrect outputs. Knowledge consistency and reasoning metrics are essential for evaluating whether an LLM's responses align with factual accuracy, maintain logical coherence, and demonstrate sound reasoning.

**Exact Match (EM)**: determines whether the generated response precisely matches the ground truth, making it a crucial metric for applications where absolute factual correctness is required [149]. This is particularly important in question-answering systems, regulatory compliance tools, and automated legal assistants, where even minor deviations in wording could alter the intended meaning. In some cases, the percentage of correct responses [150] may be used as an alternative metric, as it is conceptually similar to Exact Match (EM). This approach allows for greater flexibility in evaluation, depending on how correctness is defined, for example, by permitting paraphrased responses that preserve the original meaning or responses with minor formatting variations.

**BERTScore**: leverages deep contextual embeddings to measure semantic similarity between the generated text and reference answers, capturing meaning beyond simple word-matching [151]. Unlike Exact Match, which requires a verbatim correspondence, BERTScore assesses whether a response conveys the same meaning, even if expressed differently. This makes it highly valuable in open-ended tasks such as summarization, essay scoring, and open-domain question answering. Additionally, it is particularly useful for evaluating responses in multilingual settings, where direct translation may alter word choices but still preserve meaning.

**Logical Consistency**: refers to the ability of the model to generate outputs that follow structured reasoning principles. This ensures stable and coherent decision-making in applications such as knowledge-based reasoning, automated decision systems, and logical inference engines. Logical consistency in knowledge-integrated LLMs ensures structured reasoning by maintaining transitivity, commutativity, and negation invariance, which are crucial for stable decision-making. Quantitative metrics such as Stran, Scomm, and Sneg [152] can potentially help assess logical coherence. However, logical consistency is inherently human-centric, as it depends on how individuals interpret reasoning and make judgments, which can vary based on context, cultural background, and subjective preferences.

*5.5.8. Generation and fluency metrics*

**F1 Score:** measures the balance between precision and recall in retrieved information, making it particularly useful for evaluating RAG models in knowledge-integrated LLMs [153]. High precision ensures retrieved documents are accurate and contextually relevant, while high recall prevents missing critical knowledge needed for generating responses. Some studies report precision or recall separately based on evaluation needs—precision-focused assessments minimize false positives (e.g., medical diagnosis), whereas recall-focused evaluations ensure comprehensive retrieval (e.g., legal document discovery).

**BLEU (Bilingual Evaluation Understudy)**: evaluates n-gram overlap between generated text and reference responses, primarily assessing fluency and lexical similarity [154]. BLEU is commonly applied in machine translation and text generation, but has limitations in capturing semantic correctness and contextual relevance. A variant, *BLEU-1 (B-1)* [120], measures unigram overlap, making it particularly effective for evaluating short responses and phrase-level accuracy. While a higher B-1 score indicates better word choice alignment with reference text, it does not account for word order, coherence, or deep semantic understanding. Another refinement, *Q-BLEU-1* [120], has been designed specifically for question-answering tasks, prioritizing meaningful term matches over verbatim replication.

**ROUGE (Recall-Oriented Understudy for Gisting Evaluation)**: focuses on content coverage and information recall rather than exact lexical matches. This metric is particularly relevant for summarization tasks, where preserving the key information from a reference text is more important than replicating exact phrasing. *ROUGE* evaluates the extent to which generated text retains the essential meaning of a source document through several variations [154]. *ROUGE-N* measures n-gram recall [155], emphasizing how much of the original wording is retained. *ROUGE-2* [151], which specifically considers bigram recall, provides a better measure of coherence and phrase-level similarity between the generated and reference texts. *ROUGE-L* [120], on the other hand, leverages the longest common subsequence (LCS) between the reference and generated text, capturing sentence structure while allowing for variations in word choice. *ROUGE-W* introduces weighted scoring, placing higher importance on longer matching sequences [155], making it particularly effective for structured or factual text summarization. These methods enable ROUGE to assess knowledge-driven text generation, making it particularly suitable for domains such as legal and scientific document summarization, where content completeness and factual integrity are essential.

**Generation Time**: measures the computational efficiency of response generation, which is critical for real-time applications such as conversational AI, search engines, and customer service chatbots [156]. A reduction in generation time enhances inference speed, thereby improving the scalability and practicality of models in production environments. However, the optimization of response time must be carefully balanced against output quality and factual accuracy, as excessive emphasis on speed may increase the likelihood of hallucinations, incomplete responses, or reduced coherence. Ensuring an optimal trade-off between computational efficiency and response reliability is therefore essential for the effective deployment of knowledge-integrated language models in real-world applications.

The evaluation of knowledge-integrated Large Language Models (LLMs) relies on a diverse set of quantitative metrics that assess retrieval effectiveness, knowledge consistency, reasoning capabilities, and text generation quality.

Tables 6 and 7 provide a structured comparison of Retrieval-Augmented Generation (RAG)-based LLMs and Knowledge Graph (KG)-based LLMs, presenting key datasets, performance improvements, and essential evaluation metrics.

Based on the above results, Table 6 summarizes typical RAG-based LLM models, highlighting their reliance on external unstructured corpora for knowledge retrieval and response generation. The evaluation primarily focuses on Exact Match (EM), ROUGE, and BLEU, which assess text overlap, fluency, and factual alignment with reference responses. The results indicate that many variations of RAG models significantly enhance retrieval accuracy over baseline models, particularly in EM. However, notable performance variations are observed across different datasets, with some models exhibiting declines in EM and BLEU scores, underscoring the dataset-dependent nature of retrieval augmentation. This suggests that while RAG-based approaches effectively improve knowledge integration, their performance varies based on dataset characteristics and complexity, necessitating further optimization and adaptation for domain-specific applications.

Table 7, on the other hand, focuses on KG-based LLM models, which leverage structured knowledge representations to enhance factual accuracy and reasoning. Many recent developments have focused on integrating RAG and KG, with some hybrid approaches combining retrieval-augmented generation (RAG) and knowledge graph (KG) reasoning to leverage the strengths of both retrieval mechanisms, such as GraphRAG [151]. These hybrid models aim to enhance retrieval accuracy and reasoning capabilities while reducing hallucinations and improving factual correctness. Metrics such as *EM*, *Hits@1*, and *Recall@5* are used to evaluate these integrated models, demonstrating improvements in knowledge completeness, response fluency, and factual reliability. However, the combination of text-based retrieval and structured knowledge reasoning introduces computational trade-offs, necessitating further research on scalability and efficiency for real-world deployment.

Finally, based on Tables 1 and 2, both RAG-based, KG-based, and hybrid LLM models assess the quality of generated responses using F1 Score, Precision, and Recall, ensuring a balanced evaluation of accuracy and completeness.

**Table 6**
Performance comparison of RAG-based LLMs [157–161].

| RAG-based LLMs | Dataset | Improvement over baseline | Key metrics | Refs |
|---|---|---|---|---|
| RAG (Seq-Level) | Natural Questions (NQ) | ↑3.4% EM over DPR | EM, BLEU-1, Q-BLEU-1, ROUGE-L | [120] |
| | WebQuestions (WQ) | ↑4.1% EM over DPR | | |
| | CuratedTrec (CT) | ↑1.6% EM over DPR | | |
| | TriviaQA (TQA) | ↓0.9% EM over DPR | | |
| | Jeopardy | ↓0.4% B-1 over BART | | |
| | | ↑1.7% QB-1 over BART | | |
| | Open MS MARCO | ↑2.6% R-L over BART | | |
| | | ↑2.6% R-L over BART | | |
| ReAct | HotpotQA | ↓0.2% EM over RAG | EM, F1 | [131] |
| | | ↑8.7% F1 over RAG | | |
| | 2WikiMulti-HopQA | ↑2.2% EM over RAG | | |
| | | ↑2.8% F1 over RAG | | |
| REALM | Natural Questions (NQ) | ↑6.9% EM over ORQA | EM, Recall | [158] |
| | | ↑24.6% Recall over ORQA | | |
| GraphRAG (local) | Natural Questions (NQ) | ↓2.25% F1 over RAG | F1 Score | [151] |
| | HotpotQA | ↑1.17% F1 over RAG | | |
| | SQuALITY | ↓0.02% ROUGE-2 over RAG | ROUGE-2, BERTScore | |
| | | ↑7.04% BERTScore over RAG | | |
| | QMSum | ↑0.02% ROUGE-2 over RAG | | |
| | | ↑0.06% BERTScore over RAG | | |
| T-RAG | Customized dataset (1,614 QA pairs) | ↑16.2% Perc. over RAG | Perc. of correct responses, NIH Test | [150] |
| | | ↑23.07% NIH (k=10) over RAG | | |
| DPR-RAG | Natural Questions (NQ) | ↑8.2% EM over QRQA | EM | [159] |
| | TriviaQA (TQA) | ↑11.8% EM over QRQA | | |
| | WebQuestions (WQ) | ↑6% EM over QRQA | | |
| MoE-RAG | Google BIG-Bench | ↑0.07% F1 over RAG | F1 | [160] |
| FUNNEL-RAG | Natural Questions (NQ) | ↑3.93% recall over RAG | Recall | [161] |
| | TriviaQA (TQA) | ↑0.99% recall over RAG | | |

### 5.6. Case studies: Practical applications of LLM-knowledge base integration

This section explores real-world case studies that demonstrate the practical applications of integrating Large Language Models (LLMs) with knowledge bases. The cases discussed include FinAgent, a multimodal foundation agent designed for financial trading; the Unified Medical Language System, which facilitates the integration of medical knowledge for improved healthcare outcomes; Codex LLMs, which enhance programming capabilities by connecting code generation with vast knowledge bases; and BloombergGPT, a specialized LLM built for financial data analysis and decision-making. These case studies highlight the diverse ways in which LLM-Knowledge Base integration is applied across various industries, showcasing the potential for enhanced accuracy, decision-making, and efficiency.

#### 5.6.1. FinAgent: A multimodal foundation agent for financial trading

FinAgent gathers, processes, and analyzes diverse financial data sources, including news, stock prices, and financial reports—through a RAG-like pipeline that enhances its market intelligence capabilities. By integrating historical market trends and real-time data, FinAgent develops a comprehensive understanding of market conditions, allowing it to make data-driven and context-aware trading decisions. Additionally, it incorporates established trading strategies and expert insights as augmented tools, enabling it to combine traditional financial knowledge with its AI-driven analytical capabilities for more robust and explainable decision-making [51,98].

**Table 7**
Performance comparison of KG-based LLMs.

| KG-based LLMs | Dataset | Improvement over baseline | Key metrics | Refs |
|---|---|---|---|---|
| KGAgent | CWQ | ↑10.1% Hits@1 over GPT-4 | Hits@1, F1 | [135] |
| | | ↑18.7% F1 over GPT-4 | | |
| | WebQSP | ↑16.6% Hits@1 over GPT-4 | | |
| | | ↑19.9% F1 over GPT-4 | | |
| KAG | HotpotQA | ↑23.7% recall@5 over RAG | Exact Match (EM), F1, Recall@5 | [137] |
| | | ↑18.5% F1 over RAG | | |
| | | ↑19.1% EM over RAG | | |
| | 2WikiMulti-HopQA | ↑6.2% recall@5 over RAG | | |
| | | ↑32.9% F1 over RAG | | |
| | | ↑34.4% EM over RAG | | |
| | MuSiQue | ↑16.5% recall@5 over RAG | | |
| | | ↑22.3% F1 over RAG | | |
| | | ↑21.2% EM over RAG | | |
| ToG | CWQ | ↑21.2% accuracy over CoT | Accuracy | [130] |
| | WebQSP | ↑14.2% accuracy over CoT | | |
| KG-RAG4SM | MIMIC | ↓0.13 s generation time over GPT4o-mini | F1, generation time | [156] |
| | | same F1 with GPT4o-mini | | |
| | Synthea | ↑6.45% F1 over GPT4o-mini | | |
| | | ↓0.09 s generation time over GPT4o-mini | | |
| | CMS | ↑1.67% F1 over GPT4o-mini | | |
| | | ↓0.01 s generation time over GPT4o-mini | | |
| | EMED | ↑0.08% F1 over GPT4o-mini | | |
| | | ↓0.06 s generation time over GPT4o-mini | | |
| SAC-KG | OIE2016 | ↑4.3% F1 over PIVE | F1, AUC | [138] |
| | | ↑2.1% AUC over PIVE | | |
| | WEB | ↑6.8% F1 over PIVE | | |
| | | ↑9.7% AUC over PIVE | | |
| | NYT | ↑5.3% F1 over PIVE | | |
| | | ↑9.4% AUC over PIVE | | |
| | PENN | ↑5.1% F1 over PIVE | | |
| | | ↑8.9% AUC over PIVE | | |

To enhance trading adaptability, FinAgent employs a task-specific prompt generator that tailors prompts to different trading scenarios and market conditions. This ensures that the LLM outputs are aligned with actionable trading strategies, improving its ability to respond to dynamic financial environments. By iteratively learning from both immediate market conditions and historical performance, FinAgent enhances its trading precision and adaptability [98].

Between 2022 and 2024, FinAgent was tested on real financial datasets, with its performance evaluated against 12 baseline trading methods, including rule-based strategies, machine learning and deep learning models, reinforcement learning algorithms, and LLM-based trading systems [98]. The evaluation dataset included five major tech stocks (AAPL, AMZN, GOOGL, MSFT, TSLA) and one cryptocurrency (ETHUSD). Results showed that FinAgent consistently outperformed baseline methods across key financial performance metrics, particularly in terms of profitability. Its success is attributed to its multimodal market intelligence, augmented tools, diversified memory retrieval, and dual-level reflection, all of which contribute to its ability to adapt and optimize trading strategies dynamically.

FinAgent' architecture and functionality align with core RAG Principles: which are external knowledge integration, contextualized decision-making, leverages memory and retrieval and reasoning and reflective capabilities [95,98,107]. However, it differs from a traditional RAG, it handles multimodal data including text, numerical and visual information while rag deals with text. It integrates the RAG Principle in reinforcement Learning enabling dynamic adaptation of the market not typically found in a standard RAG. Sources lack insights on retrieval methods such as keyword-based, and semantic similarity which is discussed in detail in context. Overall, it integrates an rag-approach to enhance decision-making in financial trading.

Overall, FinAgent incorporates a RAG-inspired approach to financial trading, enhancing decision-making through knowledge integration, multimodal analysis, and adaptive reinforcement learning. Its combination of LLMs, market intelligence, and memory-enhanced learning positions it as a powerful tool for real-time financial strategy optimization.

### 5.6.2. The unified medical language system

The Unified Medical Language System (UMLS) is a comprehensive and influential resource developed by the U.S. National Library of Medicine to integrate a vast array of medical vocabularies and standards. Positioned at the foundation of multi-level medical graphs, UMLS plays a crucial role in enhancing interoperability across healthcare systems while facilitating medical diagnosis and treatment recommendations [52,162].

Strategically, UMLS serves as a foundational data source within multi-level medical graphs, anchoring medical entities and their relationships in a structured framework. By enhancing entity linking and bridging knowledge gaps, UMLS provides a robust foundation for Retrieval-Augmented Generation (RAG) applications in the medical domain [105,148,162]. Additionally, its standardized language structure supports the development of AI-driven tools for medical diagnosis, treatment planning, and decision support.

UMLS comprises three core components that contribute to its role in medical knowledge representation and integration [52,163]:

- Meta-Thesaurus: Integrates over 2 million medical vocabularies (e.g., SNOMED CT, ICD-10) encompassing nearly one million medical concepts from more than 60 biomedical vocabulary families, alongside 12 million interrelationships, ensuring a unified knowledge structure.
- Semantic Network: Organizes medical concepts hierarchically, defining relationships between them to enhance reasoning and contextual understanding in AI-driven applications.
- Lexicon: Provides linguistic insights to facilitate natural language processing (NLP) tasks within the medical domain.

Recent studies have demonstrated that integrating UMLS knowledge pathways into LLMs results in significant improvements in diagnostic accuracy and medical question answering. Research findings indicate that grounding LLM-generated responses in UMLS-enhanced knowledge graphs leads to performance gains in terms of factual correctness, completeness, and contextual relevance [162].

Moreover, studies have highlighted UMLS-augmented LLM frameworks as effective in enhancing factual accuracy and reducing hallucinations in medical AI systems. These improvements are particularly relevant for clinical decision support systems, where the quality and reliability of AI-generated outputs are critical for patient safety and medical efficacy [52].

Despite its benefits, the integration of UMLS with LLMs presents several challenges. Knowledge base integration remains a technical hurdle, as seamless retrieval and utilization of UMLS knowledge within LLM-generated outputs require sophisticated methodologies. Regular updates to UMLS are essential to incorporate new medical discoveries, evolving guidelines, and updated clinical practices, necessitating ongoing maintenance to ensure the accuracy and relevance of the knowledge base. The vast size of UMLS introduces computational overhead, requiring optimized retrieval mechanisms to balance efficiency and accuracy in AI-driven applications. Additionally, UMLS follows a symbolic representation of knowledge, whereas LLMs operate on sub-symbolic representations, creating a gap that must be bridged to enable seamless integration. Bias mitigation is another critical concern, as there is limited research on how biases in UMLS data are addressed in LLM-based applications, raising concerns about fairness and accuracy in AI-driven medical decision-making [52,162].

In conclusion, UMLS represents a critical resource in the development of sophisticated and reliable medical AI systems by enhancing interoperability, factual accuracy, and clinical decision support. However, addressing challenges related to knowledge integration, scalability, and bias mitigation will be essential to maximize its potential. Continued research and development efforts will play a vital role in optimizing the integration of UMLS with LLMs, paving the way for more reliable, explainable, and clinically relevant AI applications in healthcare.

### 5.6.3. Codex LLMs

Codex, a variant of GPT-3 developed by OpenAI, is a large language model (LLM) known for its capabilities in code generation, test assertion generation, program repair, and documentation [36,164,165]. Unlike other GPT models, Codex is specifically trained on publicly available data from GitHub, programming documentation, and coding forums, enhancing its proficiency in understanding and generating code. While it follows the transformer-based architecture of its predecessors, its specialized training dataset makes it particularly effective for software development tasks.

One of Codex's key advancements is its integration with retrieval mechanisms, forming Retrieval-Augmented Generation (RAG) systems. This approach enhances generative capabilities by enabling Codex to access and retrieve relevant external knowledge, improving the accuracy, contextuality, and relevance of generated code [36]. By incorporating retrieval-based augmentation, Codex can leverage programming documentation, API references, and contextual search methods, reducing hallucinations and improving code quality.

A significant implementation of Codex is its integration into GitHub Copilot, a code assistant that operates within integrated development environments (IDEs) such as Visual Studio Code and JetBrains [165, 166]. Copilot is facilitated by an intermediary plugin that captures user inputs, analyzes contextual information (from comments, function signatures, or existing code), and generates relevant code suggestions. The architectural components of Codex in this system include an API layer, context engine, model fine-tuning, and security mechanisms. The data flow begins with input capture within the IDE, which is then parsed for contextual analysis. The parsed input is transmitted to the Codex model for code generation, and the generated code is returned to the GitHub Copilot plugin, where it is presented for user review, selection, and feedback [166,167].

Codex has been integrated into multiple RAG-based systems, each employing unique retrieval techniques. The EPR system, introduced in 2021, integrates Codex with GPT-3, J, and Neo, utilizing demonstration-based retrieval and inference-time reasoning [148]. In 2022, OpenBook was developed, incorporating GOPHER LLM for question answering and fact verification, combining semantic encoding and symbolic reasoning to improve retrieval efficiency [148]. Another model, DSP, released in the same year, uses GPT-3.5 for open-domain and multiple-choice question answering, leveraging ColBERTv2 for retrieval augmentation [148]. In 2023, IRCoT was introduced, integrating GPT-3 and Flan-T5 for question answering while utilizing BM25 for search optimization [36,148]. Additionally, CodeLLaMA has been explored as a specialized model for code-related tasks, demonstrating overlaps with Codex's capabilities and applications [148].

Despite its versatility, Codex encounters several challenges in real-world implementations. One of the primary concerns is its limited contextual understanding in large codebases, which affects its ability to maintain coherence and logical consistency across extensive projects [36]. The computational demands of Codex contribute to latency and performance bottlenecks, making real-time inference costly. Additionally, security concerns related to code vulnerabilities and biases in training data pose risks to software integrity and ethical programming practices [165]. Codex inherits biases present in open-source code repositories, potentially propagating insecure coding patterns or reinforcing suboptimal design choices. Furthermore, an over-reliance on AI-generated code without a deep understanding of underlying programming principles can lead to misinterpretations and erroneous implementations, underscoring the need for responsible AI-assisted coding practices.

The evaluation of Codex-generated code remains an area of active research. While platforms such as Leetcode provide standardized datasets for assessing AI-generated code solutions, existing benchmarks — including noise robustness, negative rejection, information integration, and counterfactual robustness — exhibit limitations in evaluating LLM-generated code [90]. To ensure reliable assessments of code-generating LLMs, future research should focus on developing comprehensive evaluation frameworks that account for logical correctness, security vulnerabilities, maintainability, and efficiency in real-world software development.

### 5.6.4. BloombergGPT

BloombergGPT is recognized as the first mixed-domain financial LLM (FinLLM), built upon the BLOOM model and designed specifically for financial applications [12,104]. With 50 billion parameters, BloombergGPT aligns with the domain-specific pre-training paradigm, combining a general corpus of 345 billion tokens from books, articles, websites, and code with a larger financial dataset known as FinPile, comprising 363 billion tokens. FinPile includes data from diverse financial sources, such as web content, news articles, regulatory filings, press releases, and Bloomberg's proprietary datasets. This extensive financial pre-training makes BloombergGPT well-suited for specialized financial analysis and decision-making tasks.

As one of the most advanced financial LLMs, BloombergGPT represents a major milestone in the evolution of AI-driven financial applications. The performance of BloombergGPT has been evaluated across multiple task categories, including five benchmark financial NLP tasks, 12 internal Bloomberg tasks, and 42 general-purpose NLP tasks. One notable result is its 43% exact match (EM) accuracy score on the ConvFinQA dataset, which, while demonstrating strong financial domain adaptation, remains below the performance of a general crowd (47%) and significantly lower than GPT-4 with zero-shot prompting (69%–76%) [12]. Despite this, BloombergGPT's design and training data suggest its applicability across a variety of financial use cases, including sentiment analysis, risk assessment, portfolio management, financial forecasting, fraud detection, and regulatory compliance. These capabilities position BloombergGPT as a specialized tool for financial professionals, enhancing insights and streamlining analytical processes.

BloombergGPT also faces a range of challenges and limitations. Data quality and bias remain key concerns, as biases in financial data can lead to skewed predictions or reinforce pre-existing market inequalities. The lack of model explainability poses risks in financial decision-making, where transparency is crucial for regulatory compliance and investor trust. Additionally, there are ethical considerations regarding the potential for market manipulation, the creation of unfair advantages for select investors, and the displacement of human financial professionals. Addressing these issues requires responsible AI development, enhanced regulatory oversight, and transparent governance frameworks to ensure that FinLLMs are deployed in an ethical and accountable manner [12,104].

### 5.7. Challenges in LLM-knowledge integration

#### 5.7.1. Knowledge drift and long-term reliability

A fundamental challenge in integrating LLMs with knowledge bases is ensuring data reliability and sustainability over time. Knowledge bases, whether structured (e.g., knowledge graphs) or unstructured (e.g., retrieval-augmented sources), require continuous updates to reflect new discoveries, evolving policies, and domain-specific advancements. However, without efficient mechanisms for incremental learning and adaptive updates, LLMs risk retrieving outdated, inaccurate, or inconsistent information, leading to knowledge drift. This issue is particularly critical in high-stakes applications, such as law, medicine, and finance, where factual accuracy directly influences decision-making and regulatory compliance. Additionally, many knowledge bases are statically curated, making it challenging to incorporate real-time updates, dynamic corrections, or evolving contextual nuances. Addressing these limitations requires hybrid retrieval and knowledge management techniques, including automated knowledge synchronization, federated updates, and dynamic fact-checking mechanisms to ensure that LLMs continuously access relevant, current, and contextually accurate information.

#### 5.7.2. Complexity in knowledge representation and utilization

Effectively representing domain-specific knowledge in a format that LLMs can accurately interpret remains a significant challenge in knowledge integration. Many knowledge bases (KBs) and knowledge graphs (KGs) contain complex hierarchical structures, logical dependencies, and specialized terminologies that do not seamlessly align with LLM embeddings. If representations are too rigid, LLMs struggle to generalize across queries, whereas overly abstract formats may lead to hallucinated relationships and factual inconsistencies. This challenge is particularly pronounced in medical AI, legal reasoning, and financial analytics, where structured knowledge follows precise ontological frameworks and domain-specific constraints. Furthermore, effective knowledge integration requires LLMs to process both explicit symbolic representations and implicit contextual relationships, which often exceeds their native capabilities. To bridge this gap, hybrid symbolic-neural approaches, including graph-based attention mechanisms, ontology-guided prompt engineering, and knowledge-enhanced embeddings, are essential for improving LLMs' ability to retrieve, interpret, and reason over structured domain-specific knowledge accurately.

#### 5.7.3. Scalability and efficient knowledge retrieval

As the size and complexity of knowledge bases grow, managing efficient retrieval and integration with LLMs becomes increasingly challenging. Knowledge bases, especially large-scale domain-specific repositories, contain vast amounts of structured and unstructured data, making it difficult for retrieval systems to filter, rank, and process the most relevant information in real-time. This challenge is exacerbated in retrieval-augmented generation (RAG) models, where retrieving low-quality, noisy, or redundant data can negatively impact response accuracy and efficiency. Additionally, computational constraints limit the feasibility of querying large-scale knowledge bases, particularly in real-time applications where latency must be minimized. Potential solutions such as more efficient multi-stage retrieval pipelines, memory-efficient indexing, and distributed retrieval architectures can help improve scalability by ensuring that LLMs retrieve, process, and generate knowledge-enhanced responses efficiently without excessive resource consumption.

### 5.7.4. Evaluation and benchmarking limitations

Defining standardized benchmarks for LLM-knowledge base (KB) integration remains a significant challenge due to the diverse applications, empirical nature, and human-centric evaluation criteria involved in many real-world use cases. Unlike traditional NLP tasks, which rely on well-defined quantitative metrics, evaluating knowledge-enhanced LLMs necessitates measuring both retrieval effectiveness and reasoning depth, which are inherently difficult to quantify. Many applications, such as medical diagnosis, legal analysis, and financial forecasting, involve complex decision-making processes where empirical validation and expert human judgment play a crucial role, making objective benchmarking inconsistent across domains.

Furthermore, as discussed in Section 5.5, evaluating knowledge-integrated LLMs requires assessing both knowledge consistency and reasoning quality, which remain challenging to measure comprehensively. While existing evaluation methods incorporate knowledge reasoning and consistency metrics, such as Stran, Scomm, and Sneg for logical consistency, these metrics do not fully capture the depth of knowledge integration, adaptability to evolving information, or real-world applicability in empirical and human-centric applications. Addressing these limitations requires hybrid benchmarking approaches, integrating automated performance metrics with expert-driven qualitative evaluations, alongside the development of domain-specific benchmarks that account for the contextual nuances and dynamic nature of knowledge-intensive applications.

### 5.8. Recommendations for future development and implementation

This section outlines key recommendations for improving the modularity, adaptability, and efficiency of LLM systems. These recommendations focus on enhancing pipeline modularity, implementing iterative refinement processes, optimizing integration strategies, and strategically selecting between open-source and proprietary models. By adopting these recommendations, organizations can enhance the adaptability, scalability, and efficiency of LLMs, ensuring they remain relevant and effective in dynamic, real-world applications.

### 5.8.1. Modularity in pipelines

A modular architecture is essential for enhancing the flexibility, maintainability, and scalability of LLM-knowledge systems. Given the challenge of knowledge drift and long-term reliability, modular pipelines allow individual components, such as retrieval, reasoning, and generation modules, to be independently updated and optimized. The adoption of self-improving frameworks, such as DsPy, facilitates the seamless integration of incremental updates, advanced retrieval mechanisms, and evolving domain-specific knowledge, ensuring that the system remains accurate and contextually relevant over time.

Furthermore, the modular approach supports the continuous incorporation of emerging technologies and optimization strategies without necessitating a complete system overhaul. By enabling the interchangeability and independent enhancement of key components, such as retrievers and generators, this approach ensures that LLMs remain adaptable to specific domain requirements. This adaptability enhances the system's applicability across diverse industries, including healthcare, finance, and legal AI, where domain-specific knowledge updates and retrieval efficiency are critical to maintaining high-performance and trustworthiness.

Additionally, modularity facilitates the development of hybrid benchmarking frameworks, which are necessary to address the limitations of standardized LLM evaluation methods. Given the lack of consistent benchmarks for knowledge integration, modular benchmarking pipelines enable adaptive evaluation strategies by incorporating both automated metrics and expert-driven qualitative assessments. This approach ensures that evaluation methodologies remain scalable, flexible, and aligned with evolving domain-specific needs, allowing organizations to refine LLM performance assessment over time.

### 5.8.2. Iterative process

To mitigate knowledge drift and ensure long-term model relevance, LLMs must operate within an iterative learning framework rather than as static systems. Given the scalability challenges in integrating vast and evolving knowledge bases, an iterative process enables continuous refinement of retrieval mechanisms, reasoning capabilities, and factual consistency. Regular updates to knowledge bases, fine-tuning on fresh datasets, and optimization of retrieval techniques are necessary to align LLMs with emerging trends and domain-specific advancements.

This iterative approach also ensures that LLMs remain agile and responsive to industry changes, addressing shifting regulatory standards, ethical considerations, and advancements in reasoning capabilities. Particularly in highly dynamic fields such as healthcare, finance, and legal AI, iterative updates help models maintain factual accuracy, minimize outdated information, and improve real-time decision-making.

### 5.8.3. Optimization integration strategies

Maximizing the efficiency, trustworthiness, and adaptability of LLM-knowledge integration requires addressing complex knowledge representation and scalability in retrieval systems. A key optimization strategy is enhancing retrieval mechanisms to improve efficiency and knowledge selection, ensuring that LLMs retrieve high-quality, domain-relevant information while minimizing noise and redundancy. Implementing hybrid retrieval strategies, such as combining sparse and dense retrieval techniques with multi-stage ranking, allows LLMs to process and utilize structured and unstructured knowledge sources more effectively.

Beyond retrieval, improving text generation and contextual integration ensures that LLMs incorporate retrieved knowledge more accurately and coherently. Current models often struggle with factual inconsistencies or hallucinations when integrating external knowledge. Optimizing input-layer fusion, intermediate-layer integration, and retrieval-conditioned decoding enhances contextual coherence and domain-specific accuracy, allowing LLMs to better reason over structured knowledge representations.

Additionally, domain-specific adaptation is essential for improving structured reasoning in fields such as medicine, finance, and legal AI, where knowledge follows precise ontological structures. Fine-tuning models on specialized datasets ensures more accurate interpretation and application of domain-specific knowledge, reducing reliance on generic pre-trained data.

### 5.8.4. Strategic selection between open-source and proprietary models

The choice between open-source and proprietary LLMs is crucial in balancing cost, performance, customization, and regulatory compliance. Proprietary models often deliver high performance, but they come with significant financial costs, limited transparency, and potential vendor lock-in. In contrast, open-source models offer greater control, adaptability, and domain-specific fine-tuning capabilities, making them preferable for knowledge-intensive applications that require data privacy and regulatory compliance.

Fine-tuning smaller open-source models on specialized medical, legal, or financial datasets can yield competitive performance compared to proprietary alternatives while ensuring data sovereignty. Additionally, hybrid retrieval-augmented generation (RAG) pipelines can bridge the gap between open-source and proprietary models, enhancing retrieval precision and reasoning quality while maintaining cost-effectiveness. In highly regulated sectors such as healthcare and finance, open-source models provide an advantage by ensuring full oversight of data processing and compliance with privacy regulations.

##### 5.8.5. Complementary AI approaches for LLM enhancement

While this survey primarily focuses on the integration of Large Language Models (LLMs) with knowledge-based methods, recent research has explored several alternative AI augmentation approaches to enhance model efficiency, reasoning capabilities, and factual accuracy beyond traditional knowledge integration. One such approach is Symbolic AI, which has been integrated into LLMs to improve logical reasoning and interpretability by combining rule-based knowledge representation with neural network-based learning [168]. Similarly, Neuroscience-Inspired AI leverages cognitive science principles, incorporating architectures such as hierarchical temporal memory and spiking neural networks to enhance energy efficiency and mimic human-like reasoning processes [169]. Another promising direction is multi-agent collaboration, where multiple specialized AI agents interact and coordinate to solve complex tasks [170], improving task decomposition, modular problem-solving, and adaptability in real-time environments. Additionally, Self-Optimizing [171] and Reinforcement-Based Learning [172] employs adaptive feedback mechanisms that allow models to iteratively refine their performance and decision-making capabilities based on continuous learning and user interactions. While these approaches differ from knowledge-based integration, they offer complementary advantages. Future research may explore hybrid models that synergistically combine these techniques with knowledge-integrated LLMs, further expanding AI's ability to generate reliable, efficient, and contextually aware outputs.

## 6. Conclusion

This paper has explored the integration of Large Language Models (LLMs) with knowledge bases, beginning with an overview of LLMs to understand their key capabilities and real-world applications. It summarizes the challenges faced in implementing LLMs for real-world scenarios. We also examined existing solutions and innovations aimed at overcoming these challenges, emphasizing the value of hybrid approaches that leverage the strengths of both LLMs and knowledge bases.

Subsequently, we conducted a comprehensive analysis of AI enhancement through the integration of LLMs with knowledge-based systems. By reviewing the current state of integration, key techniques, evaluation metrics, and representative case studies, we identified critical benefits, challenges, and future directions. Our findings demonstrate that this integration enhances data contextualization, improves model accuracy, and facilitates more reliable knowledge retrieval across various domains. However, challenges remain, particularly in dynamic knowledge management and model adaptability. Future research should focus on refining integration techniques, optimizing retrieval processes, and ensuring that knowledge bases remain current and relevant.

In conclusion, integrating LLMs with knowledge bases offers significant potential to advance AI technology and improve its application across diverse sectors. The continued evolution of this field promises to result in more intelligent, accurate, and context-aware AI systems, with substantial benefits for various industries and organizations.

## CRediT authorship contribution statement

**Wenli Yang:** Writing – review & editing, Writing – original draft, Visualization, Validation, Supervision, Resources, Methodology, Data curation, Conceptualization. **Lilian Some:** Writing – original draft, Resources, Methodology, Data curation. **Michael Bain:** Writing – review & editing, Supervision, Resources. **Byeong Kang:** Writing – review & editing, Supervision, Methodology, Conceptualization.

## Declaration of competing interest

The authors declare that they have no known competing financial interests or personal relationships that could have appeared to influence the work reported in this paper.

## Data availability

No data was used for the research described in the article.

## References

[1] Shervin Minaee, Tomas Mikolov, Narjes Nikzad, Meysam Chenaghlu, Richard Socher, Xavier Amatriain, Jianfeng Gao, Large language models: A survey, 2024.

[2] Zichong Wang, Zhibo Chu, Thang Viet Doan, Shiwen Ni, Min Yang, Wenbin Zhang, History, development, and principles of large language models: an introductory survey, 2024.

[3] Sanjay Kukreja, Tarun Kumar, Amit Purohit, Abhijit Dasgupta, Debashis Guha, A literature survey on open source large language models, 2024.

[4] Teo Susnjak, Peter Hwang, Napoleon H. Reyes, Andre L.C. Barczak, Timothy R. McIntosh, Surangika Ranathunga, Automating research synthesis with domain-specific large language model fine-tuning, 2024.

[5] Timothy R. McIntosh, Teo Susnjak, Tong Liu, Paul Watters, Malka N. Halgamuge, The inadequacy of reinforcement learning from human feedback-radicalizing large language models via semantic vulnerabilities, 2023.

[6] Nuraini Sulaiman, Farizal Hamzah, Evaluation of transfer learning and adaptability in large language models with the glue benchmark, 2024.

[7] Mohaimenul Azam Khan Raiaan, Md. Saddam Hossain Mukta, Kaniz Fatema, Nur Mohammad Fahad, Sadman Sakib, Most Marufatul Jannat Mim, Jubaer Ahmad, Mohammed Eunus Ali, Sami Azam, A review on large language models: Architectures, applications, taxonomies, open issues and challenges, IEEE Access 12 (2024) 26839–26874.

[8] Amanda Kau, Xuzeng He, Aishwarya Nambissan, Aland Astudillo, Hui Yin, Amir Aryani, Combining knowledge graphs and large language models, 2024, arXiv:2407.06564 [cs].

[9] Huy Quoc To, Ming Liu, Guangyan Huang, Towards efficient large language models for scientific text: A review, 2024, arXiv:2408.10729 [cs].

[10] Muhammad Usman Hadi, Qasem Al Tashi, Rizwan Qureshi, Abbas Shah, Amgad Muneer, Muhammad Irfan, Anas Zafar, Muhammad Bilal Shaikh, Naveed Akhtar, Jia Wu, Seyedali Mirjalili, A survey on large language models: Applications, challenges, limitations, and practical usage, 2023.

[11] Wayne Xin Zhao, Kun Zhou, Junyi Li, Tianyi Tang, Xiaolei Wang, Yupeng Hou, Yingqian Min, Beichen Zhang, Junjie Zhang, Zican Dong, et al., A survey of large language models, 2023, arXiv preprint arXiv:2303.18223, 1(2).

[12] Jean Lee, Nicholas Stevens, Soyeon Caren Han, Minseok Song, A survey of large language model in finance (FinLLMs), 2024, arXiv:2402.02315 [cs, q-fin].

[13] Jacob Devlin, Ming-Wei Chang, Kenton Lee, Kristina Toutanova, BERT: Pre-training of deep bidirectional transformers for language understanding, 2018.

[14] Enkelejda Kasneci, Kathrin Sessler, Stefan Küchemann, Maria Bannert, Daryna Dementieva, Frank Fischer, Urs Gasser, Georg Groh, Stephan Günnemann, Eyke Hüllermeier, Stephan Krusche, Gitta Kutyniok, Tilman Michaeli, Claudia Nerdel, Jürgen Pfeffer, Oleksandra Poquet, Michael Sailer, Albrecht Schmidt, Tina Seidel, Matthias Stadler, Jochen Weller, Jochen Kuhn, Gjergji Kasneci, ChatGPT for good? On opportunities and challenges of large language models for education, Learn. Individ. Differ. 103 (2023) 102274.

[15] Quanjun Zhang, Chunrong Fang, Yang Xie, YuXiang Ma, Weisong Sun, Yun Yang, Zhenyu Chen, A systematic literature review on large language models for automated program repair, 2024, arXiv:2405.01466 [cs].

[16] Xiao Bi, Deli Chen, Guanting Chen, Shanhuang Chen, Damai Dai, Chengqi Deng, Honghui Ding, Kai Dong, Qiushi Du, Zhe Fu, et al., DeepSeek LLM: Scaling open-source language models with longtermism, 2024, arXiv preprint arXiv:2401.02954.

[17] Daya Guo, Qihao Zhu, Dejian Yang, Zhenda Xie, Kai Dong, Wentao Zhang, Guanting Chen, Xiao Bi, Yu Wu, Y.K. Li, et al., Deepseek-coder: When the large language model meets programming–the rise of code intelligence, 2024, arXiv preprint arXiv:2401.14196.

[18] Damai Dai, Chengqi Deng, Chenggang Zhao, R.X. Xu, Huazuo Gao, Deli Chen, Jiashi Li, Wangding Zeng, Xingkai Yu, Yu Wu, et al., DeepSeekMoE: Towards ultimate expert specialization in mixture-of-experts language models, 2024, arXiv preprint arXiv:2401.06066.

[19] Abhimanyu Dubey, Abhinav Jauhri, Abhinav Pandey, Abhishek Kadian, Ahmad Al-Dahle, Aiesha Letman, Akhil Mathur, Alan Schelten, Amy Yang, Angela Fan, et al., The llama 3 herd of models, 2024, arXiv preprint arXiv:2407.21783.

[20] Qinggang Zhang, Shengyuan Chen, Yuanchen Bei, Zheng Yuan, Huachi Zhou, Zijin Hong, Junnan Dong, Hao Chen, Yi Chang, Xiao Huang, A survey of graph retrieval-augmented generation for customized large language models, 2025, arXiv preprint arXiv:2501.13958.

[21] Han Zhang, Langshi Zhou, Hanfang Yang, Learning to retrieve and reason on knowledge graph through active self-reflection, 2025, arXiv preprint arXiv:2502.14932.

[22] Lei Huang, Weijiang Yu, Weitao Ma, Weihong Zhong, Zhangyin Feng, Haotian Wang, Qianglong Chen, Weihua Peng, Xiaocheng Feng, Bing Qin, et al., A survey on hallucination in large language models: Principles, taxonomy, challenges, and open questions, ACM Trans. Inf. Syst. 43 (2) (2025) 1–55.

[23] Tianyang Zhong, Zhengliang Liu, Yi Pan, Yutong Zhang, Yifan Zhou, Shizhe Liang, Zihao Wu, Yanjun Lyu, Peng Shu, Xiaowei Yu, et al., Evaluation of OpenAI o1: Opportunities and challenges of AGI, 2024, arXiv preprint arXiv: 2409.18486.

[24] Erik Keanius, Domain adaptation of LLMs: A study of content generation, RAG, and fine-tuning, 2024.

[25] Chengrui Wang, Qingqing Long, Meng Xiao, Xunxin Cai, Chengjun Wu, Zhen Meng, Xuezhi Wang, Yuanchun Zhou, BioRAG: A RAG-LLM framework for biological question reasoning, 2024, arXiv:2408.01107 [cs].

[26] Desta Haileselassie Hagos, Rick Battle, Danda B. Rawat, Recent advances in generative AI and large language models: Current status, challenges, and perspectives, 2024, arXiv:2407.14962 [cs].

[27] Benjamin Warner, Antoine Chaffin, Benjamin Clavié, Orion Weller, Oskar Hallström, Said Taghadouini, Alexis Gallagher, Raja Biswas, Faisal Ladhak, Tom Aarsen, et al., Smarter, better, faster, longer: A modern bidirectional encoder for fast, memory efficient, and long context finetuning and inference, 2024, arXiv preprint arXiv:2412.13663.

[28] Tyna Eloundou, Sam Manning, Pamela Mishkin, Daniel Rock, Gpts are gpts: An early look at the labor market impact potential of large language models, 2023.

[29] Xiaonan Li, Changtai Zhu, Linyang Li, Zhangyue Yin, Tianxiang Sun, Xipeng Qiu, LLatrieval: LLM-verified retrieval for verifiable generation, 2024, arXiv: 2311.07838 [cs].

[30] Mike Lewis, Yinhan Liu, Naman Goyal, Marjan Ghazvininejad, Abdelrahman Mohamed, Omer Levy, Ves Stoyanov, Luke Zettlemoyer, Bart: Denoising sequence-to-sequence pre-training for natural language generation, translation, and comprehension, 2019, arXiv preprint arXiv:1910.13461.

[31] Shuming Ma, Li Dong, Shaohan Huang, Dongdong Zhang, Alexandre Muzio, Saksham Singhal, Hany Hassan Awadalla, Xia Song, Furu Wei, Deltalm: Encoder–decoder pre-training for language generation and translation by augmenting pretrained multilingual encoders, 2021, arXiv preprint arXiv:2106. 13736.

[32] Pranjal Kumar, Large language models (llms): survey, technical frameworks, and future challenges, Artif. Intell. Rev. 57 (10) (2024) 260.

[33] Mohaimenul Azam Khan Raiaan, Md. Saddam Hossain Mukta, Kaniz Fatema, Nur Mohammad Fahad, Sadman Sakib, Most Marufatul Jannat Mim, Jubaer Ahmad, Mohammed Eunus Ali, Sami Azam, A review on large language models: Architectures, applications, taxonomies, open issues and challenges, IEEE Access 12 (2024) 26839–26874.

[34] N. Kitaev, et al., Reformer: The efficient transformer, 2020.

[35] Siyu Ding, Junyuan Shang, Shuohuan Wang, Yu Sun, Hao Tian, Hua Wu, Haifeng Wang, Ernie-doc: A retrospective long-document modeling transformer, 2020, arXiv preprint arXiv:2012.15688.

[36] Yunfan Gao, Yun Xiong, Xinyu Gao, Kangxiang Jia, Jinliu Pan, Yuxi Bi, Yi Dai, Jiawei Sun, Meng Wang, Haofen Wang, Retrieval-augmented generation for large language models: A survey, 2024, arXiv:2312.10997 [cs].

[37] Nourhan Ibrahim, Samar Aboulela, Ahmed Ibrahim, Rasha Kashef, A survey on augmenting knowledge graphs (kgs) with large language models (llms): models, evaluation metrics, benchmarks, and challenges, Discov. Artif. Intell. 4 (1) (2024) 76.

[38] Amin Beheshti, Empowering generative ai with knowledge base 4.0: Towards linking analytical, cognitive, and generative intelligence, 2023.

[39] Andrea Matarazzo, Riccardo Torlone, A survey on large language models with some insights on their capabilities and limitations, 2025, arXiv preprint arXiv: 2501.04040.

[40] Taiyu Zhang, Xuesong Zhang, Robbe Cools, Adalberto L. Simeone, Focus agent: LLM-powered virtual focus group, 2024, arXiv:2409.01907 [cs].

[41] Wenqi Fan, Yujuan Ding, Liangbo Ning, Shijie Wang, Hengyun Li, Dawei Yin, Tat-Seng Chua, Qing Li, A survey on RAG meeting LLMs: Towards retrieval-augmented large language models, 2024, arXiv:2405.06211 [cs].

[42] Lilian Some, Wenli Yang, Michael Bain, Byeong Kang, A comprehensive survey on integrating large language models with knowledge-based methods, 2025, arXiv preprint arXiv:2501.13947.

[43] Yingqing He, Zhaoyang Liu, Jingye Chen, Zeyue Tian, Hongyu Liu, Xiaowei Chi, Runtao Liu, Ruibin Yuan, Yazhou Yao, Wenhai Wang, Jifeng Dai, Yong Zhang, Wei Xue, Qifeng Liu, Yike Guo, Qifeng Chen, LLMs meet multimodal generation and editing: A survey, 2024, arXiv:2405.19334 [cs].

[44] Ruiyao Xu, Kaize Ding, Large language models for anomaly and out-of-distribution detection: A survey, 2024, arXiv:2409.01980 [cs].

[45] Linhao Luo, Yuan-Fang Li, Gholamreza Haffari, Shirui Pan, Reasoning on graphs: Faithful and interpretable large language model reasoning, 2024, arXiv: 2310.01061 [cs].

[46] Sander Schulhoff, Michael Ilie, Nishant Balepur, Konstantine Kahadze, Amanda Liu, Chenglei Si, Yinheng Li, Aayush Gupta, HyoJung Han, Sevien Schulhoff, et al., The prompt report: A systematic survey of prompting techniques, 2024, arXiv preprint arXiv:2406.06608.

[47] Richard Fang, Rohan Bindu, Akul Gupta, Daniel Kang, LLM agents can autonomously exploit one-day vulnerabilities, 2024, arXiv:2404.08144 [cs].

[48] Marissa Radensky, Daniel S. Weld, Joseph Chee Chang, Pao Siangliulue, Jonathan Bragg, Let's get to the point: Llm-supported planning, drafting, and revising of research-paper blog posts, 2024, arXiv preprint arXiv:2406.10370.

[49] Xiaofei Sun, Xiaoya Li, Shengyu Zhang, Shuhe Wang, Fei Wu, Jiwei Li, Tianwei Zhang, Guoyin Wang, Sentiment analysis through llm negotiations, 2023, arXiv preprint arXiv:2311.01876.

[50] Rajvardhan Patil, Venkat Gudivada, A review of current trends, techniques, and challenges in large language models (llms), Appl. Sci. 14 (5) (2024) 2074.

[51] Yuqi Nie, Yaxuan Kong, Xiaowen Dong, John M. Mulvey, H. Vincent Poor, Qingsong Wen, Stefan Zohren, A survey of large language models for financial applications: Progress, prospects and challenges, 2024, arXiv:2406.11903 [q-fin].

[52] Rui Yang, Edison Marrese-Taylor, Yuhe Ke, Lechao Cheng, Qingyu Chen, Irene Li, Integrating umls knowledge into large language models for medical question answering, 2023.

[53] Zabir Al Nazi, Wei Peng, Large language models in healthcare and medical domain: A review, in: Informatics, Vol. 11, MDPI, 2024, p. 57.

[54] Arun James Thirunavukarasu, Darren Shu Jeng Ting, Kabilan Elangovan, Laura Gutierrez, Ting Fang Tan, Daniel Shu Wei Ting, Large language models in medicine, Nature Med. 29 (8) (2023) 1930–1940.

[55] Frank Fagan, A view of how language models will transform law, Tenn. Law Rev. 92 (2024) 1.

[56] Yinheng Li, Shaofei Wang, Han Ding, Hang Chen, Large language models in finance: A survey, in: Proceedings of the Fourth ACM International Conference on AI in Finance, 2023, pp. 374–382.

[57] Yanjun Gao, Ruizhe Li, Emma Croxford, John Caskey, Brian W. Patterson, Matthew Churpek, Timothy Miller, Dmitriy Dligach, Majid Afshar, Leveraging medical knowledge graphs into large language models for diagnosis prediction: Design and application study, JMIR AI 4 (2025) e58670.

[58] Huanghai Liu, Quzhe Huang, Qingjing Chen, Yiran Hu, Jiayu Ma, Yun Liu, Weixing Shen, Yansong Feng, Jurex-4e: Juridical expert-annotated four-element knowledge base for legal reasoning, 2025, arXiv:2502.17166.

[59] Jean Lee, Nicholas Stevens, Soyeon Caren Han, Large language models in finance (finllms), Neural Comput. Appl. (2025) 1–15.

[60] Zibin Zheng, Kaiwen Ning, Qingyuan Zhong, Jiachi Chen, Wenqing Chen, Lianghong Guo, Weicheng Wang, Yanlin Wang, Towards an understanding of large language models in software engineering tasks, Empir. Softw. Eng. 30 (2) (2025) 50.

[61] Jingzhi Gong, Vardan Voskanyan, Paul Brookes, Fan Wu, Wei Jie, Jie Xu, Rafail Giavrimis, Mike Basios, Leslie Kanthan, Zheng Wang, Language models for code optimization: Survey, challenges and future directions, 2025, arXiv preprint arXiv:2501.01277.

[62] Jiayang Wu, Wensheng Gan, Zefeng Chen, Shicheng Wan, S. Yu Philip, Multimodal large language models: A survey, in: 2023 IEEE International Conference on Big Data, BigData, IEEE, 2023, pp. 2247–2256.

[63] Dawei Huang, Chuan Yan, Qing Li, Xiaojiang Peng, From large language models to large multimodal models: A literature review, Appl. Sci. 14 (12) (2024) 5068.

[64] Zijing Liang, Yanjie Xu, Yifan Hong, Penghui Shang, Qi Wang, Qiang Fu, Ke Liu, A survey of multimodel large language models, in: Proceedings of the 3rd International Conference on Computer, Artificial Intelligence and Control Engineering, 2024, pp. 405–409.

[65] Karthik Meduri, Hari Gonaygunta, Geeta Sandeep Nadella, Priyanka Pramod Pawar, Deepak Kumar, Adaptive intelligence: Gpt-powered language models for dynamic responses to emerging healthcare challenges, IJARCCE 13 (2024) 104–109.

[66] Iker García-Ferrero, Rodrigo Agerri, Aitziber Atutxa Salazar, Elena Cabrio, Iker de la Iglesia, Alberto Lavelli, Bernardo Magnini, Benjamin Molinet, Johana Ramirez-Romero, German Rigau, et al., Medical mt5: an open-source multilingual text-to-text llm for the medical domain, 2024, arXiv preprint arXiv:2404.07613.

[67] Angela Fan, Beliz Gokkaya, Mark Harman, Mitya Lyubarskiy, Shubho Sengupta, Shin Yoo, Jie M. Zhang, Large language models for software engineering: Survey and open problems, in: 2023 IEEE/ACM International Conference on Software Engineering: Future of Software Engineering, ICSE-FoSE, IEEE, 2023, pp. 31–53.

[68] Jiaqi Wang, Enze Shi, Huawen Hu, Chong Ma, Yiheng Liu, Xuhui Wang, Yincheng Yao, Xuan Liu, Bao Ge, Shu Zhang, Large language models for robotics: Opportunities, challenges, and perspectives, J. Autom. Intell. (2024).

[69] Sonal Samrajngh, Thiago Fraga-Silva, Youssef Oualil, Christophe Van Gysel, Synthetic query generation using large language models for virtual assistants, in: Proceedings of the 47th International ACM SIGIR Conference on Research and Development in Information Retrieval, 2024, pp. 2837–2841.

[70] Defne Yigci, Merve Eryilmaz, Ail K. Yetisen, Savas Tasoglu, Aydogan Ozcan, Large language model-based chatbots in higher education, Adv. Intell. Syst. (2024) 2400429.

[71] Ting-Chi Chang, Yu-Jou Chen, Sheng Hung, Ning-Hsuan Chang, Chih-Hao Ku, Szu-Yin Lin, Shih-Yi Chien, A review on shaping chatbot personalities via large language models, 2025.

[72] Devon Myers, Rami Mohawesh, Venkata Ishwarya Chellaboina, Anantha Lakshmi Sathvik, Praveen Venkatesh, Yi-Hui Ho, Hanna Henshaw, Muna Al-hawawreh, David Berdik, Yaser Jararweh, Foundation and large language models: fundamentals, challenges, opportunities, and social impacts, Clust. Comput. 27 (1) (2024) 1–26.

[73] Anqi Wang, Zhizhuo Yin, Yulu Hu, Yuanyuan Mao, Pan Hui, Exploring the potential of large language models in artistic creation: Collaboration and reflection on creative programming, 2024, arXiv preprint arXiv:2402.09750.

[74] Hung-Fu Chang, Tong Li, A framework for collaborating a large language model tool in brainstorming for triggering creative thoughts, Think. Ski. Creativity (2025) 101755.

[75] Robin Qiu, Large language models: from entertainment to solutions, Digit. Transform. Soc. 3 (2) (2024) 125–126.

[76] Karthik Soman, Peter W. Rose, John H. Morris, Rabia E. Akbas, Brett Smith, Braian Peetoom, Catalina Villouta-Reyes, Gabriel Cerono, Yongmei Shi, Angela Rizk-Jackson, Sharat Israni, Charlotte A. Nelson, Sui Huang, Sergio E. Baranzini, Biomedical knowledge graph-optimized prompt generation for large language models, Bioinformatics 40 (9) (2024) btae560.

[77] Long Ouyang, Jeff Wu, Xu Jiang, Diogo Almeida, Carroll L. Wainwright, Pamela Mishkin, Chong Zhang, Sandhini Agarwal, Katarina Slama, Alex Ray, John Schulman, Jacob Hilton, Fraser Kelton, Luke Miller, Maddie Simens, Amanda Askell, Peter Welinder, Paul Christiano, Jan Leike, Ryan Lowe, Training language models to follow instructions with human feedback, 2022.

[78] Xin Cheng, Di Luo, Xiuying Chen, Lemao Liu, Dongyan Zhao, Rui Yan, Lift yourself up: Retrieval-augmented text generation with self memory, 2023.

[79] YunHe Su, Zhengyang Lu, Junhui Liu, Ke Pang, Haoran Dai, Sa Liu Yuxin Jia, Lujia Ge, Jing-min Yang, Applications of large models in medicine, 2025, arXiv preprint arXiv:2502.17132.

[80] Hongzhi Zhang, M. Omair Shafiq, Triple-aware reasoning: A retrieval-augmented generationapproach for enhancing question-answering tasks with-knowledge graphs and large language models, 2024, https://caiac.pubpub.org/pub/bytcy6lo.

[81] Shengjie Ma, Chengjin Xu, Xuhui Jiang, Muzhi Li, Huaren Qu, Jian Guo, Think-on-graph 2.0: Deep and interpretable large language model reasoning with knowledge graph-guided retrieval, 2024, arXiv:2407.10805 [cs].

[82] Yinghao Zhu, Changyu Ren, Shiyun Xie, Shukai Liu, Hangyuan Ji, Zixiang Wang, Tao Sun, Long He, Zhoujun Li, Xi Zhu, Chengwei Pan, REALM: RAG-driven enhancement of multimodal electronic health records analysis via large language models, 2024, arXiv:2402.07016 [cs].

[83] Xinyang Hu, Fengzhuo Zhang, Siyu Chen, Zhuoran Yang, Unveiling the statistical foundations of chain-of-thought prompting methods, 2024, arXiv:2408.14511 [cs, stat, math].

[84] Alaa Abd-alrazaq, Rawan AlSaad, Dari Alhuwail, Arfan Ahmed, Padraig Mark Healy, Syed Latifi, Sarah Aziz, Rafat Damseh, Sadam Alabed Alrazak, Javaid Sheikh, Large language models in medical education: Opportunities, challenges, and future directions, JMIR Med. Educ. 9 (2023) e48291.

[85] Yun-Tong Yang, Hong-Gang Luo, Topological or not? A unified pattern description in the one-dimensional anisotropic quantum XY model with a transverse field, 2023, arXiv:2302.13866 [cond-mat].

[86] Tailin Liang, John Glossner, Lei Wang, Shaobo Shi, Xiaotong Zhang, Pruning and quantization for deep neural network acceleration: A survey, 2021.

[87] Xiaokai Wei, Shen Wang, Dejiao Zhang, Parminder Bhatia, Andrew Arnold, Knowledge enhanced pretrained language models: A comprehensive survey, 2021, arXiv:2110.08455 [cs].

[88] Sébastien Bubeck, Varun Chandrasekaran, Ronen Eldan, Johannes Gehrke, Eric Horvitz, Ece Kamar, Peter Lee, Yin Tat Lee, Yuanzhi Li, Scott Lundberg, Harsha Nori, Hamid Palangi, Marco Tulio Ribeiro, Yi Zhang, Sparks of artificial general intelligence: Early experiments with gpt-4, 2023.

[89] Jordan Hoffmann, Sebastian Borgeaud, Arthur Mensch, Elena Buchatskaya, Trevor Cai, Eliza Rutherford, Diego de Las Casas, Lisa Anne Hendricks, Johannes Welbl, Aidan Clark, et al., Training compute-optimal large language models, 2022, arXiv preprint arXiv:2203.15556.

[90] Jiawei Chen, Hongyu Lin, Xianpei Han, Le Sun, Benchmarking large language models in retrieval-augmented generation, Proc. AAAI Conf. Artif. Intell. 38 (16) (2024) 17754–17762.

[91] Nicholas Matsumoto, Jay Moran, Hyunjun Choi, Miguel E. Hernandez, Mythreye Venkatesan, Paul Wang, Jason H. Moore, KRAGEN: a knowledge graph-enhanced RAG framework for biomedical problem solving using large language models, Bioinformatics 40 (6) (2024) btae353.

[92] Bodong Chen, Xinran Zhu, Fernando Díaz Del Castillo H., Integrating generative AI in knowledge building, Comput. Educ.: Artif. Intell. 5 (2023) 100184, MAG ID: 4388599367.

[93] Muhammad, Zeeshan Nawaz, Mahmoud Fahmy, Navigating challenges and technical debt in large language models deployment, in: Proceedings of the 4th Workshop on Machine Learning and Systems, 2024, pp. 192–199.

[94] Xin Su, Tiep Le, Steven Bethard, Phillip Howard, Semi-structured chain-of-thought: Integrating multiple sources of knowledge for improved language model reasoning, 2024, arXiv:2311.08505 [cs].

[95] Tilmann Bruckhaus, Rag does not work for enterprises, 2024.

[96] Thomas Wolf, Lysandre Debut, Victor Sanh, Julien Chaumond, Clement Delangue, Anthony Moi, Pierric Cistac, Tim Rault, Rémi Louf, Morgan Funtowicz, Joe Davison, Sam Shleifer, Patrick von Platen, Clara Ma, Yacine Jernite, Julien Plu, Canwen Xu, Teven Le Scao, Sylvain Gugger, Mariama Drame, Quentin Lhoest, Alexander M. Rush, HuggingFace's transformers: State-of-the-art natural language processing, 2019.

[97] Miao Zheng, Hao Liang, Fan Yang, Haoze Sun, Tianpeng Li, Lingchu Xiong, Yan Zhang, Youzhen Wu, Kun Li, Yanjun Shen, Mingan Lin, Tao Zhang, Guosheng Dong, Yujing Qiao, Kun Fang, Weipeng Chen, Bin Cui, Wentao Zhang, Zenan Zhou, Pas: Data-efficient plug-and-play prompt augmentation system, 2024, arXiv:2407.06027 [cs].

[98] Wentao Zhang, Lingxuan Zhao, Haochong Xia, Shuo Sun, Jiaze Sun, Molei Qin, Xinyi Li, Yuqing Zhao, Yilei Zhao, Xinyu Cai, Longtao Zheng, Xinrun Wang, Bo An, A multimodal foundation agent for financial trading: Tool-augmented, diversified, and generalist, 2024, arXiv:2402.18485 [q-fin].

[99] Zhengbao Jiang, Frank F. Xu, Luyu Gao, Zhiqing Sun, Qian Liu, Jane Dwivedi-Yu, Yiming Yang, Jamie Callan, Graham Neubig, Active retrieval augmented generation, 2023, arXiv:2305.06983 [cs].

[100] Xijun Wang, Dongshan Ye, Chenyuan Feng, Howard H. Yang, Xiang Chen, Tony Q.S. Quek, Trustworthy image semantic communication with GenAI: Explainability, controllability, and efficiency, 2024, ARXIV_ID: 2408.03806 S2ID: 3e5473ecb44e13e6bb08b477623ab39da551943b.

[101] Fiona Fui-Hoon Nah, Ruilin Zheng, Jingyuan Cai, Keng Siau, Langtao Chen, Generative AI and ChatGPT: Applications, challenges, and AI-human collaboration, J. Inf. Technol. Case Appl. Res. 25 (3) (2023) 277–304.

[102] Ian J. Goodfellow, Jean Pouget-Abadie, Mehdi Mirza, Bing Xu, David Warde-Farley, Sherjil Ozair, Aaron Courville, Yoshua Bengio, Generative adversarial networks, 2014.

[103] D.P. Kingma, M. Welling, Auto-encoding variational bayes, 2013.

[104] Aidan Hogan, Eva Blomqvist, Michael Cochez, Claudia d'Amato, Gerard de Melo, Claudio Gutierrez, José Emilio Labra Gayo, Sabrina Kirrane, Sebastian Neumaier, Axel Polleres, Roberto Navigli, Axel-Cyrille Ngonga Ngomo, Sabbir M. Rashid, Anisa Rula, Lukas Schmelzeisen, Juan Sequeda, Steffen Staab, Antoine Zimmermann, Knowledge graphs, ACM Comput. Surv. 54 (4) (2022) 1–37, arXiv:2003.02320 [cs].

[105] Minhyul Jeong, Jiwoong Sohn, Mujeen Sung, Jaewoo Kang, Improving medical reasoning through retrieval and self-reflection with retrieval-augmented large language models, 2024, arXiv:2401.15269 [cs].

[106] Emma Yann Zhang, Adrian David Cheok, Zhigeng Pan, Jun Cai, Ying Yan, From turing to transformers: A comprehensive review and tutorial on the evolution and applications of generative transformer models, Sci. 5 (4) (2023) 46.

[107] Peizhong Gao, Ao Xie, Shaoguang Mao, Wenshan Wu, Yan Xia, Haipeng Mi, Furu Wei, Meta reasoning for large language models, 2024, arXiv:2406.11698 [cs].

[108] Dr. Kasnesis, Augmentation of large language model capabilities with knowledge graphs, 2024.

[109] Weijian Xie, Xuefeng Liang, Yuhui Liu, Kaihua Ni, Hong Cheng, Zetian Hu, Weknow-RAG: An adaptive approach for retrieval-augmented generation integrating web search and knowledge graphs, 2024, arXiv:2408.07611 [cs].

[110] Sijia Chen, Baochun Li, Di Niu, Boosting of thoughts: Trial-and-error problem solving with large language models, 2024, arXiv:2402.11140 [cs].

[111] Ilia Stepin, Muhammad Suffian, Alejandro Catalá, J. Alonso-Moral, How to build self-explaining fuzzy systems: From interpretability to explainability [AI-eXplained], 2024, S2ID: 1d5127686d01fe1e66805c10f19284f865484632.

[112] Ling Yang, Zhaochen Yu, Tianjun Zhang, Shiyi Cao, Minkai Xu, Wentao Zhang, Joseph E. Gonzalez, Bin Cui, Buffer of thoughts: Thought-augmented reasoning with large language models, 2024, arXiv:2406.04271 [cs].

[113] Majid Afshar, Yanjun Gao, Deepak Gupta, Emma Croxford, Dina Demner-Fushman, On the role of the umls in supporting diagnosis generation proposed by large language models, J. Biomed. Inform. 157 (2024) 104707.

[114] Lukas Bahr, Christoph Wehner, Judith Wewerka, José Bittencourt, Ute Schmid, Rüdiger Daub, Daub knowledge graph enhanced retrieval-augmented generation for failure mode and effects analysis, 2024, arXiv:2406.18114 [cs].

[115] Garima Agrawal, Tharindu Kumarage, Zeyad Alghamdi, Huan Liu, Can knowledge graphs reduce hallucinations in LLMs? : A survey, 2023, arXiv:2311.07914 [cs].

[116] Hasan Abu-Rasheed, Christian Weber, Madjid Fathi, Knowledge graphs as context sources for LLM-based explanations of learning recommendations, 2024, arXiv:2403.03008 [cs].

[117] Shirui Pan, Linhao Luo, Yufei Wang, Chen Chen, Jiapu Wang, Xindong Wu, Unifying large language models and knowledge graphs: A roadmap, IEEE Trans. Knowl. Data Eng. 36 (7) (2024) 3580–3599, arXiv:2306.08302 [cs].

[118] Julien Delile, Srayanta Mukherjee, Anton Van Pamel, Leonid Zhukov, Graph-based retriever captures the long tail of biomedical knowledge, 2024, arXiv:2402.12352 [cs].

[119] Paulo Finardi, Leonardo Avila, Rodrigo Castaldoni, Pedro Gengo, Celio Larcher, Marcos Piau, Pablo Costa, Vinicius Caridá, The chronicles of RAG: The retriever, the chunk and the generator, 2024, arXiv:2401.07883 [cs].

[120] Patrick Lewis, Ethan Perez, Aleksandra Piktus, Fabio Petroni, Vladimir Karpukhin, Naman Goyal, Heinrich Küttler, Mike Lewis, Wen-tau Yih, Tim Rocktäschel, et al., Retrieval-augmented generation for knowledge-intensive nlp tasks, Adv. Neural Inf. Process. Syst. 33 (2020) 9459–9474.

[121] Junde Wu, Jiayuan Zhu, Yunli Qi, Medical graph RAG: Towards safe medical large language model via graph retrieval-augmented generation, 2024, arXiv:2408.04187 [cs].

[122] Yu Wang, Shiwan Zhao, Zhihu Wang, Heyuan Huang, Ming Fan, Yubo Zhang, Zhixing Wang, Haijun Wang, Ting Liu, Strategic chain-of-thought: Guiding accurate reasoning in LLMs through strategy elicitation, 2024, arXiv:2409.03271 [cs].

[123] Jinyuan Fang, Zaiqiao Meng, Craig Macdonald, TRACE the evidence: Constructing knowledge-grounded reasoning chains for retrieval-augmented generation, 2024, arXiv:2406.11460 [cs].

[124] Davit Janezashvili, Rag at large enterprises, 2024, https://modulai.io/blog/rag-at-large-enterprises/. (Accessed 25 October 2024).

[125] Haoyu Wang, Ruirui Li, Haoming Jiang, Jinjin Tian, Zhengyang Wang, Chen Luo, Xianfeng Tang, Monica Cheng, Tuo Zhao, Jing Gao, BlendFilter: Advancing retrieval-augmented large language models via query generation blending and knowledge filtering, 2024, arXiv:2402.11129 [cs].

[126] Yujia Zhou, Zheng Liu, Jiajie Jin, Jian-Yun Nie, Zhicheng Dou, Metacognitive retrieval-augmented large language models, 2024, arXiv:2402.11626 [cs].

[127] Masoomali Fatehkia, Ji Kim Lucas, Sanjay Chawla, T-RAG: Lessons from the LLM trenches, 2024, arXiv:2402.07483 [cs].

[128] Darren Edge, Ha Trinh, Newman Cheng, Joshua Bradley, Alex Chao, Apurva Mody, Steven Truitt, Dasha Metropolitansky, Robert Osazuwa Ness, Jonathan Larson, From local to global: A graph rag approach to query-focused summarization, 2024, arXiv preprint arXiv:2404.16130.

[129] Zhenyu Li, Sunqi Fan, Yu Gu, Xiuxing Li, Zhichao Duan, Bowen Dong, Ning Liu, Jianyong Wang, Flexkbqa: A flexible llm-powered framework for few-shot knowledge base question answering, in: Proceedings of the AAAI Conference on Artificial Intelligence, Vol. 38, 2024, pp. 18608–18616.

[130] Jiashuo Sun, Chengjin Xu, Lumingyuan Tang, Saizhuo Wang, Chen Lin, Yeyun Gong, Lionel M. Ni, Heung-Yeung Shum, Jian Guo, Think-on-graph: Deep and responsible reasoning of large language model on knowledge graph, 2023, arXiv preprint arXiv:2307.07697.

[131] Shunyu Yao, Jeffrey Zhao, Dian Yu, Nan Du, Izhak Shafran, Karthik Narasimhan, Yuan Cao, ReAct: Synergizing reasoning and acting in language models, 2023, arXiv:2210.03629 [cs].

[132] Shengjie Ma, Chengjin Xu, Xuhui Jiang, Muzhi Li, Huaren Qu, Jian Guo, Think-on-graph 2.0: Deep and interpretable large language model reasoning with knowledge graph-guided retrieval, 2024, arXiv e-prints, pages arXiv–2407.

[133] Yifei Zhang, Xintao Wang, Jiaqing Liang, Sirui Xia, Lida Chen, Yanghua Xiao, Chain-of-knowledge: Integrating knowledge reasoning into large language models by learning from knowledge graphs, 2024, arXiv preprint arXiv:2407.00653.

[134] Weijian Xie, Xuefeng Liang, Yuhui Liu, Kaihua Ni, Hong Cheng, Zetian Hu, Weknow-rag: An adaptive approach for retrieval-augmented generation integrating web search and knowledge graphs, 2024, arXiv preprint arXiv:2408.07611.

[135] Jinhao Jiang, Kun Zhou, Wayne Xin Zhao, Yang Song, Chen Zhu, Hengshu Zhu, Ji-Rong Wen, Kg-agent: An efficient autonomous agent framework for complex reasoning over knowledge graph, 2024, arXiv preprint arXiv:2402.11163.

[136] Qitan Lv, Jie Wang, Hanzhu Chen, Bin Li, Yongdong Zhang, Feng Wu, Coarse-to-fine highlighting: Reducing knowledge hallucination in large language models, 2024, arXiv preprint arXiv:2410.15116.

[137] Lei Liang, Mengshu Sun, Zhengke Gui, Zhongshu Zhu, Zhouyu Jiang, Ling Zhong, Yuan Qu, Peilong Zhao, Zhongpu Bo, Jin Yang, et al., Kag: Boosting llms in professional domains via knowledge augmented generation, 2024, arXiv preprint arXiv:2409.13731.

[138] Hanzhu Chen, Xu Shen, Qitan Lv, Jie Wang, Xiaoqi Ni, Jieping Ye, Sac-kg: Exploiting large language models as skilled automatic constructors for domain knowledge graphs, 2024, arXiv:2410.02811.

[139] Jiejun Tan, Zhicheng Dou, Yutao Zhu, Peidong Guo, Kun Fang, Ji-Rong Wen, Small models, big insights: Leveraging slim proxy models to decide when and what to retrieve for LLMs, 2024, arXiv:2402.12052 [cs].

[140] Zhihong Shao, Yeyun Gong, Yelong Shen, Minlie Huang, Nan Duan, Weizhu Chen, Enhancing retrieval-augmented large language models with iterative retrieval-generation synergy, 2023, arXiv:2305.15294 [cs].

[141] Yanming Liu, Xinyue Peng, Xuhong Zhang, Weihao Liu, Jianwei Yin, Jiannan Cao, Tianyu Du, RA-ISF: Learning to answer and understand from retrieval augmentation via iterative self-feedback, 2024, arXiv:2403.06840 [cs].

[142] Zhentao Xu, Mark Jerome Cruz, Matthew Guevara, Tie Wang, Manasi Deshpande, Xiaofeng Wang, Zheng Li, Retrieval-augmented generation with knowledge graphs for customer service question answering, 2024, arXiv:2404.17723 [cs].

[143] Jialin Dong, Bahare Fatemi, Bryan Perozzi, Lin F. Yang, Anton Tsitsulin, Don't forget to connect! improving RAG with graph-based reranking, 2024, arXiv:2405.18414 [cs].

[144] Zhibin Gou, Zhihong Shao, Yeyun Gong, Yelong Shen, Yujiu Yang, Minlie Huang, Nan Duan, Weizhu Chen, ToRA: A tool-integrated reasoning agent for mathematical problem solving, 2024, arXiv:2309.17452 [cs].

[145] Taicheng Guo, Xiuying Chen, Yaqi Wang, Ruidi Chang, Shichao Pei, Nitesh V. Chawla, Olaf Wiest, Xiangliang Zhang, Large language model based multi-agents: A survey of progress and challenges, 2024.

[146] M. Karabacak, K. Margetis, Embracing large language models for medical applications: Opportunities and challenges, Cureus 15 (5) (2023) e39305.

[147] Xiaoxi Li, Jiajie Jin, Yujia Zhou, Yuyao Zhang, Peitian Zhang, Yutao Zhu, Zhicheng Dou, From matching to generation: A survey on generative information retrieval, 2024, arXiv:2404.14851 [cs].

[148] C. Zakka, R. Shad, A. Chaurasia, A.R. Dalal, J.L. Kim, M. Moor, R. Fong, C. Phillips, K. Alexander, E. Ashley, J. Boyd, K. Boyd, K. Hirsch, C. Langlotz, R. Lee, J. Melia, J. Nelson, K. Sallam, S. Tullis, M.A. Vogelsong, W. Hiesinger, Almanac - retrieval-augmented language models for clinical medicine, 2024.

[149] Taojun Hu, Xiao-Hua Zhou, Unveiling llm evaluation focused on metrics: Challenges and solutions, 2024, arXiv preprint arXiv:2404.09135.

[150] Masoomali Fatehkia, Ji Kim Lucas, Sanjay Chawla, T-rag: lessons from the llm trenches, 2024, arXiv preprint arXiv:2402.07483.

[151] Haoyu Han, Harry Shomer, Yu Wang, Yongjia Lei, Kai Guo, Zhigang Hua, Bo Long, Hui Liu, Jiliang Tang, Rag vs. graphrag: A systematic evaluation and key insights, 2025, arXiv preprint arXiv:2502.11371.

[152] Yinhong Liu, Zhijiang Guo, Tianya Liang, Ehsan Shareghi, Ivan Vulić, Nigel Collier, Aligning with logic: Measuring, evaluating and improving logical consistency in large language models, 2024, arXiv preprint arXiv:2410.02205.

[153] Hao Yu, Aoran Gan, Kai Zhang, Shiwei Tong, Qi Liu, Zhaofeng Liu, Evaluation of retrieval-augmented generation: A survey, in: CCF Conference on Big Data, Springer, 2024, pp. 102–120.

[154] Yupeng Chang, Xu Wang, Jindong Wang, Yuan Wu, Linyi Yang, Kaijie Zhu, Hao Chen, Xiaoyuan Yi, Cunxiang Wang, Yidong Wang, et al., A survey on evaluation of large language models, ACM Trans. Intell. Syst. Technol. 15 (3) (2024) 1–45.

[155] Sandeep Kumar, Arun Solanki, Rouge-ss: A new rouge variant for evaluation of text summarization, 2023, Authorea Preprints.

[156] Chuangtao Ma, Sriom Chakrabarti, Arijit Khan, Bálint Molnár, Knowledge graph-based retrieval-augmented generation for schema matching, 2025, arXiv preprint arXiv:2501.08686.

[157] Joel Ruben Antony Moniz, Soundarya Krishnan, Melis Ozyildirim, Prathamesh Saraf, Halim Cagri Ates, Yuan Zhang, Hong Yu, Realm: retrieval-augmented generation against adversarial and noisy data, 2024, arXiv:2403.20329.

[158] Yinghao Zhu, Changyu Ren, Shiyun Xie, Shukai Liu, Hangyuan Ji, Zixiang Wang, Tao Sun, Long He, Zhoujun Li, Xi Zhu, et al., Realm: Rag-driven enhancement of multimodal electronic health records analysis via large language models, 2024, arXiv:2402.07016.

[159] Benjamin Reichman, Larry Heck, Dense passage retrieval: Is it retrieving? 2024, arXiv:2402.11035.

[160] Xingyu Xiong, Mingliang Zheng, Merging mixture of experts and retrieval augmented generation for enhanced information retrieval and reasoning, 2024.

[161] Xinping Zhao, Yan Zhong, Zetian Sun, Xinshuo Hu, Zhenyu Liu, Dongfang Li, Baotian Hu, Min Zhang, Funnelrag: A coarse-to-fine progressive retrieval paradigm for rag, 2024, arXiv preprint arXiv:2410.10293.

[162] Majid Afshar, Yanjun Gao, Deepak Gupta, Emma Croxford, Dina Demner-Fushman, On the role of the umls in supporting diagnosis generation proposed by large language models, J. Biomed. Inform. 157 (2024) 104707.

[163] Olivier Bodenreider, The unified medical language system (umls): integrating biomedical terminology, Nucleic Acids Res. 32 (suppl_1) (2004) D267–D270.

[164] Tristan Coignion, Clément Quinton, Romain Rouvoy, A performance study of llm-generated code on leetcode, 2024.

[165] Junjielong Xu, Ziang Cui, Yuan Zhao, Xu Zhang, Shilin He, Pinjia He, Liqun Li, Yu Kang, Qingwei Lin, Yingnong Dang, Saravan Rajmohan, Dongmei Zhang, Unilog: Automatic logging via llm and in-context learning, 2024.

[166] Danie Smit, Hanlie Smuts, Paul Louw, Julia Pielmeier, Christina Eidelloth, The impact of github copilot on developer productivity from a software engineering body of knowledge perspective, 2024.

[167] Roberto Gozalo-Brizuela, Eduardo C. Garrido-Merchan, ChatGPT is not all you need. A state of the art review of large generative AI models, 2023.

[168] Sen Yang, Xin Li, Leyang Cui, Lidong Bing, Wai Lam, Neuro-symbolic integration brings causal and reliable reasoning proofs, 2023, arXiv preprint arXiv:2311.09802.

[169] Safoora Yousefi, Leo Betthauser, Hosein Hasanbeig, Raphaël Millière, Ida Momennejad, Decoding in-context learning: Neuroscience-inspired analysis of representations in large language models, 2023, arXiv preprint arXiv:2310.00313.

[170] Taicheng Guo, Xiuying Chen, Yaqi Wang, Ruidi Chang, Shichao Pei, Nitesh V. Chawla, Olaf Wiest, Xiangliang Zhang, Large language model based multi-agents: A survey of progress and challenges, 2024, arXiv preprint arXiv:2402.01680.

[171] Eric Zelikman, Eliana Lorch, Lester Mackey, Adam Tauman Kalai, Self-taught optimizer (stop): Recursively self-improving code generation, in: First Conference on Language Modeling, 2024.

[172] Rameez Qureshi, Naïm Es-Sebbani, Luis Galárraga, Yvette Graham, Miguel Couceiro, Zied Bouraoui, Refine-lm: Mitigating language model stereotypes via reinforcement learning, in: ECAI 2024, IOS Press, 2024, pp. 4027–4034.